\documentclass[10pt,twocolumn,letterpaper]{article}

\usepackage[pagenumbers]{cvpr}      

\usepackage[dvipsnames]{xcolor}

\usepackage[textsize=tiny]{todonotes}

\usepackage{graphicx}
\usepackage{booktabs}  
\usepackage{diagbox}
\usepackage{url}
\usepackage{multirow}
\usepackage{subcaption}

\definecolor{cvprblue}{rgb}{0.21,0.49,0.74}
\usepackage[pagebackref,breaklinks,colorlinks,allcolors=cvprblue]{hyperref}

\def\paperID{2718} 
\def\confName{ECCV}
\def\confYear{2026}

\usepackage{graphicx}
\usepackage{booktabs}
\usepackage{wrapfig}

\newif\ifdraft
\draftfalse

\usepackage{xparse}
\usepackage{xspace}
\usepackage{tikz}
\usepackage{adjustbox}
\usepackage{multirow}
\usepackage{colortbl}
\usepackage{array}
\usepackage{makecell}
\newcommand{\PreserveBackslash}[1]{\let\temp=\\#1\let\\=\temp}
\newcolumntype{C}[1]{>{\PreserveBackslash\centering}p{#1}}
\newcolumntype{R}[1]{>{\PreserveBackslash\raggedleft}p{#1}}
\newcolumntype{L}[1]{>{\PreserveBackslash\raggedright}p{#1}}

\usepackage{microtype}
\usepackage{graphicx}
\usepackage{mathtools}
\usepackage{subcaption}
\usepackage{booktabs} 

\usepackage{amssymb}
\usepackage{pifont}

\usepackage{bbold}

\usepackage{amssymb,enumitem}

\setlist[itemize]{leftmargin=*}
\setlist[enumerate]{leftmargin=*}

\newcommand*{\rej}{{\ooalign{\lower.3ex\hbox{$\sqcup$}\cr\raise.4ex\hbox{$\sqcap$}}}}

\usepackage{stackengine}

\usepackage{xcolor}

\renewcommand{\ie}{\textit{i.e.,}\@\xspace}
\renewcommand{\eg}{\textit{e.g.,}\@\xspace}

\graphicspath{{images/}}

\usepackage{booktabs,arydshln}
\makeatletter
\def\adl@drawiv#1#2#3{%
        \hskip.5\tabcolsep
        \xleaders#3{#2.5\@tempdimb #1{1}#2.5\@tempdimb}%
                #2\z@ plus1fil minus1fil\relax
        \hskip.5\tabcolsep}
\newcommand{\cdashlinelr}[1]{%
  \noalign{\vskip\aboverulesep
           \global\let\@dashdrawstore\adl@draw
           \global\let\adl@draw\adl@drawiv}
  \cdashline{#1}
  \noalign{\global\let\adl@draw\@dashdrawstore
           \vskip\belowrulesep}}
\makeatother

\newcommand{\nlp}[1]{}

\newcolumntype{x}[1]{>{\centering\arraybackslash\hspace{0pt}}p{#1}}

\newcommand{\ours}{MultiMem\@\xspace}
\newcommand{\clip}{CLIPMem\@\xspace}
\newcommand{\avt}{AVT-CLIP\@\xspace}
\newcommand{\avit}{AVIT-CLIP\@\xspace}
\newcommand{\dejavu}{déjà vu\@\xspace}
\newcommand{\unit}{UnitMem\@\xspace}

\newcommand{\bimodal}{bi-modal\xspace}

\usepackage{amsmath}

\usepackage{amsmath,amsfonts,bm}

\def\Secref#1{Section~\ref{#1}}

\def\eqref#1{equation~\ref{#1}}

\def\1{\bm{1}}

\DeclareMathAlphabet{\mathsfit}{\encodingdefault}{\sfdefault}{m}{sl}
\SetMathAlphabet{\mathsfit}{bold}{\encodingdefault}{\sfdefault}{bx}{n}

\newcommand{\metric}{CMC}

\usepackage{makecell}
\usepackage[flushleft]{threeparttable}

\newcounter{mycounter} 

\newcommand{\blfootnote}[1]{%
  \begingroup
  \renewcommand{\thefootnote}{}\footnotetext{#1}%
  \endgroup
}
\title{MultiMem: Measuring and Mitigating Memorization\\ in Multi-Modal Contrastive Learning} 

\author{%
Wenhao Wang, Franziska Boenisch, Michael Backes, Adam Dziedzic\\
{\normalsize \hspace{0em}\texttt{\{wenhao.wang, boenisch, backes, adam.dziedzic\}@cispa.de}}\\
CISPA Helmholtz Center for Information Security
}

\begin{document}

\maketitle
\blfootnote{Accepted at the 19th European Conference on Computer Vision (ECCV), 2026.}

\begin{abstract}
\label{sec:abstract}
Memorization in machine learning models enables high performance on rare in-distribution samples by capturing their atypical patterns. However, it also causes harmful retention of noise and outliers, degrading generalization. While memorization has been extensively studied in both supervised and self-supervised learning in the vision domain, it remains unexplored in multi-modal contrastive learning. We address this gap by introducing MultiMem, the first metric designed to quantify memorization in multi-modal contrastive learning. Through our systematic analysis, we demonstrate that cross-modal semantic misalignment has the strongest influence on memorization, with text being the dominant modality driving memorization, followed by video, image, and audio. We show that targeted augmentations applied across all modalities effectively reduce memorization as measured by our MultiMem metric and improve model performance. Overall, this work establishes the first framework for measuring and mitigating memorization in multi-modal contrastive learning, preventing harmful data retention and contributing to higher-performing models.

\end{abstract}

\section{Introduction}\label{sec:intro}


Multi-modal contrastive learning aims to jointly process and align data from diverse modalities such as images, text, audio, and video. This paradigm has demonstrated high performance across a wide range of tasks, including image captioning~\cite{liu2021cptr, hu2022scaling, zhou2023style}, visual question answering~\cite{song2022clip, yang2022zero, wang2023vqa}, zero-shot classification~\cite{radford2021learning, zhang2022tip, mu2022slip}, and cross-modal retrieval~\cite{wen2021cookie, baldrati2022conditioned, baldrati2022effective}. These successes underscore the benefits of integrating heterogeneous modalities, which lead to improved generalization and more robust semantic representations. However, it remains unclear to what extent memorization contributes to the observed improvements, motivating the need to better understand the role of memorization in multi-modal contrastive learning. Previous studies have shown that in both supervised learning (SL)~\cite{feldman2020does,feldman2020neural} and self-supervised learning (SSL)~\cite{wang2024localizing}, model's memorization of training data points is essential for generalization. In the vision domain, it has been observed that models tend to memorize outliers in the training set, which correspond to mislabeled samples in SL~\cite{bartlett2020benign, feldman2020does} and atypical examples in SSL~\cite{wang2024memorization}. A similar study of memorization in multi-modal contrastive learning is lacking. 

Multi-modal learning introduces unique challenges not seen in uni-modal settings~\cite{liang2022mind, li2024multimodal, liang2024foundations}, such as inconsistencies between modalities, varying noise levels, and the lack of explicit modality-alignment. These factors limit the direct transfer of insights from uni-modal studies to multi-modal learning. Importantly, most existing definitions of memorization are designed for specific modalities and tasks, for example, based on label prediction in SL~\cite{feldman2020does} or augmentation matching in SSL~\cite{wang2024memorization} for the vision domain. Such definitions do not generalize well to the multi-modal setting: Measuring memorization within individual modalities or among a limited subset of all the modalities used to train a given multi-modal model, fails to capture multi-modal memorization \textit{faithfully}. Therefore, a new definition tailored to multi-modal contrastive learning is needed.

There are only two studies on measuring memorization for the \bimodal image-text models like CLIP~\cite{radford2021learning}, namely \textit{\dejavu} memorization~\cite{jayaraman2024d} and \textit{\clip}~\cite{wang2025captured}.
The \textit{\dejavu} method measures memorization by masking one modality and testing whether the model can recover the other, while \clip measures memorization by comparing alignment between model pairs trained with or without a given sample (\ie an image-text pair). However, these methods are not directly applicable to multi-modal contrastive learning which involves additional modalities, such as video and audio.

To address these limitations, we propose \textbf{\ours}: \textbf{a general-purpose metric for measuring memorization in multi-modal contrastive learning}, which is designed for any number and type of modalities involved.
Specifically, \ours builds on the leave-one-out framework~\cite{feldman2020does,ye2023leave,wang2024memorization} for measuring memorization and compares the outputs of a pair of multi-modal models trained with and without a given multi-modal sample on this particular sample.

Through our extensive empirical study on multiple contrastively trained multi-modal models such as \textbf{AudioCLIP}~\cite{guzhov2022audioclip} (including \textbf{A}udio, \textbf{I}mage, and \textbf{T}ext modalities), \textbf{\avt} (a custom-built tri-modal model with \textbf{A}udio, \textbf{V}ideo, and \textbf{T}ext modalities, introduced in \Secref{sec:model_dataset}), and \textbf{\avit} (a custom-built quad-modal model with \textbf{A}udio, \textbf{V}ideo, \textbf{I}mage, and \textbf{T}ext modalities, introduced in \Secref{sec:model_dataset}), we observe that: (1) Global memorization, measured across all modalities, behaves differently from memorization observed within any subset of modalities and to capture the full extent of a model's memorization, we need to assess memorization jointly across all its modalities.
(2)~The most memorized samples are not simply mis-captioned, as reported for the most memorized samples in text-image models~\cite{wang2025captured}, but semantically misaligned across all modalities: the information provided by different modalities not only contradict each other but is mostly semantically \textit{not related}. 
(3) Multi-modal models increasingly memorize cross-modal patterns, rather than primarily text as observed in bi-modal models~\cite{wang2025captured}. This suggests that multi-modal models' memorization behavior is closer aligned with the one of SSL models~\cite{wang2024memorization} which were shown to memorize pattern, rather than SL models which memorize labels~\cite{feldman2020does}. 

Based on the above observations, we explore \textit{different strategies to mitigate memorization and improve generalization} in multi-modal models.
\textbf{In-training:} we \textit{actively measure memorization during multi-modal training} to identify the top-memorized samples. At a given stage of training, we re-group the top-memorized samples into new batches and continuously apply noise-based augmentations (only) to them in subsequent training steps. 
\textbf{Post-training:} we \textit{identify highly memorized samples post-training} and then fine-tune the model on the remaining training samples. 
Our experiments show that both approaches \textit{\textbf{substantially reduce memorization}} by up to \textbf{20\%}, while increasing model performance up to \textbf{8\%} for retrieval, \textbf{10\%} for zero-shot, and \textbf{4\%} for downstream classification tasks.
In summary: 
\begin{itemize}
    \item We propose \textit{\ours}, a metric that measures memorization by comparing the outputs of a pair of multi-modal models trained with and without a given multi-modal sample on this particular sample
    \item Our analysis with \ours \textit{demonstrates the key differences between multi-modal models with more than two modalities and previously studied bi-modal or uni-modal models}, including distinct memorization patterns, a shift from label-driven to pattern-driven memorization behavior, and a stronger influence of cross-modal semantic inconsistency on memorization.
    \item Based on our findings, we propose \textit{two approaches to mitigate memorization and improve generalization in multi-modal models}: either during training or after training. Our extensive experiments show that both methods effectively reduce the model’s memorization level and lead to substantial improvements in performance.
\end{itemize}

\section{Background and Related Work}

\paragraph{Multi-Modal Contrastive Learning.}\label{sec:multimodal}
Contrastive Language-Image Pretraining (\textbf{CLIP})~\cite{radford2021learning} proposed a joint training framework for image and text modalities, where text and image representations are aligned using the InfoNCE loss~\cite{oord2018representation}.
CLIP enables strong zero-shot image classification by using language prompts as class representations, and supports accurate image-text retrieval through a shared representation space.
\textbf{VideoCLIP}~\cite{xu2021videoclip}
extends CLIP to the video domain by substituting the original image encoder with a transformer-based video encoder. 
The video is first uniformly sampled into a fixed number of frames, which are individually processed by a frozen convolutional neural network (CNN) with a trainable multi-layer perceptron (MLP) to obtain frame-level representations.
These frame-level representations are then encoded as tokens by a transformer and aggregated by applying average pooling to obtain the representation for the entire video.
This modification enables the model to process temporal-visual information while retaining a contrastive training framework. VideoCLIP shows strong performance in zero-shot video understanding tasks, such as action recognition and video-text retrieval, without task-specific supervision.
\textbf{AudioCLIP}~\cite{guzhov2022audioclip} extends CLIP to the audio domain by introducing an additional audio encoder based on an ESResNeXt convolutional network~\cite{guzhov2021esresnet}. The audio input is first converted into a log-mel spectrogram and then projected into a shared representation space, where it is aligned with image and text representations. The model is trained using the same contrastive objective as CLIP, encouraging alignment between semantically related audio, image, and text samples, while pushing apart mismatched pairs. As the model incorporates one more modality than CLIP, the training loss is computed as the sum of the pairwise InfoNCE losses (\ie \textbf{A}udio-\textbf{I}mage, \textbf{A}udio-\textbf{T}ext, and \textbf{I}mage-\textbf{T}ext).
AudioCLIP performs well in zero-shot audio classification and cross-modal retrieval tasks, outperforming previous methods that rely on task-specific supervision.



\vspace{1em}
\noindent \textbf{Memorization in SL \& SSL.}
Although memorization has been associated with potential risks of sensitive information leakage, multiple studies~\cite{feldman2020does, sadrtdinov2021memorization, wang2024memorization} in supervised learning (SL) and self-supervised learning (SSL) suggest that memorization also plays an important role in the generalization of models.
They also show that both SL and SSL models tend to memorize atypical samples during training, while the nature of these atypical samples varies. For example, in SL, memorization often correlates with mislabeled or noisy samples, while in SSL, it tends to occur on images with rare or distinctive visual patterns~\cite{meehan2023ssl,wang2024localizing}. 

A common definition of label memorization in SL is the leave-one-out style definition by~\cite{feldman2020neural}:
\begin{equation}
\label{equ:aligment_multi}
\vspace{-1em}
\begin{split}
\textbf{Mem}\left( \mathcal{A}, S, i\right) = \underset{f\leftarrow \mathcal{A}(S)}{\textbf{Pr}}\left[ f\left( x_i\right) = y_i\right] \\
- \underset{g\leftarrow \mathcal{A}(S^{\setminus i})}{\textbf{Pr}}\left[ g\left( x_i\right) = y_i\right]
\text{,} 
\end{split}
\vspace{-1em}
\end{equation}
where $\mathcal{A}$ is a training algorithm (here for models $f$ and $g$) and $S^{\setminus i}$ represents train set $S$ without the data point $(x_i,y_i)$.
In this definition, a data point is considered memorized if the model's prediction of the point's ground truth training label changes significantly based on whether the point was or was not used to train the model.

For SSL, where no labels are available, \cite{wang2024memorization} proposed a new metric:
\begin{equation}\label{eq:memdef}
    \begin{split}
    \textbf{SSLMem}(S,Aug,i) =  \underset{ 
    \substack{g \sim \mathcal{A}(S^ {\setminus {i}})\\ {x_i}',{x_i}''\sim \text{Aug}({x_i})}}{\mathbb{E}}
    \left[ d\left( g\left({x_i}' \right), g\left({x_i}'' \right)\right)\right]\\
    - \underset{
   \substack{ f \sim \mathcal{A} (S)\\ {x_i}',{x_i}''\sim \text{Aug}({x_i})}}
   {\mathbb{E}} 
    \left[ d\left( f\left({x_i}' \right), f\left({x_i}'' \right)\right)\right]
    \text{,} 
    \end{split}
\end{equation}
where $d(\cdot , \cdot)$ represents a distance function, commonly the $\ell_2$ 
distance, and $\text{Aug}(\cdot)$ is the augmentation sets used during training. Here, the expectation captures the \textit{alignment} of the respective models as the expected Euclidean distance 
between their representations of two augmentations of the same sample. 
Then, SSLMem is computed as the \textit{alignment difference} between the model pairs trained with and without that sample. 
\textit{Both metrics are limited to uni-modal models, making them unsuitable for evaluating memorization effects in multi-modal settings.}

\textbf{Memorization in Bi-modal Contrastive Models.}
\cite{jayaraman2024d} proposed a retrieval-based evaluation protocol to detect \textit{\dejavu} memorization in vision-language models. Their method assesses whether a model trained on one data subset can retrieve objects that appeared only in a different, held-out subset, thereby indicating unintended memorization across disjoint training sets.
However, these approaches only offer qualitative evaluations of whether memorization occurs, without providing a quantitative measurement of how strongly a model memorizes specific data points across all modalities.

The only existing method for \textit{quantifying} memorization across the text and vision modalities is \clip~\cite{wang2025captured}.
It computes the difference in alignment scores between a pair of CLIP-style models, trained with and without this data point. Formally, \clip is defined as:
\begin{equation}
\text{\textbf{\clip}}(I, T) = \mathcal{A}_{\text{align}}(f, I, T) - \mathcal{A}_{\text{align}}(g, I, T) \text{,}\quad 
\end{equation}
where $f$ and $g$ are CLIP models trained on datasets with and without a data point $(I,T)$, consisting of an \textbf{I}mage and \textbf{T}ext (caption). The alignment score 
$\mathcal{A}_{\text{align}}$ is computed as the cosine similarity between image and text representations, corrected by subtracting similarities to unrelated text and image samples $(i,t)$ that were not used in training $f$ or $g$.

\begin{equation}
\begin{split}
\mathcal{A}_{\text{align}}(f, I, T) = \underset{(I',T')  \sim \text{Aug}(I,T)}{\mathbb{E}} \left[\text{sim}(f_{\text{img}}(I'), f_{\text{txt}}(T'))\right] \\
- \underset{t}{\mathbb{E}} \left[ \text{sim}(f_{\text{img}}(I), f_{\text{txt}}(t)) \right] - \underset{i}{\mathbb{E}} \left[ \text{sim}(f_{\text{img}}(i), f_{\text{txt}}(T)) \right] \text{,}
\end{split}
\end{equation}
where $(i,t)$ is a set of randomly chosen image and text testing samples that were not used in training $f$ or $g$. 
CLIPMem indicates that samples where caption and corresponding image do not align well (so-called \textbf{mis-captioned} samples) are most memorized.
While \clip is limited to measuring bi-modal memorization between \textbf{I}mage and \textbf{T}ext modalities, our \ours metric generalizes it to capture memorization among \textit{many more diverse modalities}.

\section{Our MultiMem Metric}


The memorization metrics discussed in the previous section are tailored to uni-modal or bi-modal models and do not account for global interactions among all modalities in a multi-modal setting. We demonstrate empirically in \Cref{fig:multi_vs_clip} that these approaches are insufficient for faithfully capturing memorization in multi-modal models. Therefore, we introduce our \ours metric, which is specifically designed to quantify \textit{global} memorization across all modalities.



We build \ours on the leave-one-out framework~\cite{feldman2020does, wang2024memorization, wang2025captured} to measure memorization with respect to a pair of models, $f$ and $g$. Here, model $f$ is trained on the full training set $S$, while model $g$ is trained on the subset $S^{\setminus i}$ where the multi-modal sample $x_i$ was removed.

Given the complex interactions among modalities in multi-modal contrastive learning, we first require a proxy that quantifies the quality of the model’s representations of an input data point $x_i$ which can be compared between models.
Since the learning objective in multi-modal contrastive learning is to make the representations returned by the model on the different modalities of the same sample as consistent as possible, and the representations of unrelated examples as far as possible, we consider cross‑modal consistency ($\metric$) on the sample itself vs. on unrelated samples as a proxy.
We calculate the cross-modal consistency as follows: 
Given a model with $n$ different modalities, we define the representation space $\Phi$ as:
\begin{equation}
\label{equ:phi}
\begin{split}
\Phi_{x_i} = 
\begin{bmatrix}
\hat{\phi}_1\\
\hat{\phi}_2 \\
\vdots\\
\hat{\phi}_n
\end{bmatrix}
\in \mathbb{R}^{n \times d}\text{,}
\end{split}
\end{equation}
where $\phi_j \in \mathbb{R}^d$ is the representation of $x_i$ in the $j$-th modality, and $\hat{\phi}_j$ denotes its $\ell_2$-normalized version.

Then we define cross-modal consistency $\metric(i, H)$ on data point $x_i$ (with respect to a held-out set $H$) as:
\begin{equation}
\label{equ:phi}
\begin{split}
\metric
(i, H) = 
\frac{1}{2}\underset{(\cdot)'\sim \text{Aug}}{\mathbb{E}}\mathbf{1}_{n}^{\top}
\left( {\Phi_{x_{i}^{'}}}{\Phi^{\top}_{x_{i}^{'}}} \right)
\mathbf{1}_{n} \\
- \frac{1}{2}\underset{^{\ \ \ h \in H}_{(\cdot)'\sim \text{Aug}}}{\mathbb{E}}\mathbf{1}_{n}^{\top}
\left( {\Phi_{x_{i}^{'}}}{\Phi^{\top}_{{h}'}} \right)
\mathbf{1}_{n}\text{,}
\end{split}
\end{equation}
where $\mathbf{1}_n$ denotes an all-ones vector of length $n$ and $(\cdot)'\sim \text{Aug}$ indicates that we compute an expectation over representations computed on different random augmentations of the samples, increasing stability of the metric. The held-out set are randomly picked from non-trained samples in the validation set of training datasets.

The first term of the score captures the similarities across all modality pairs within the representation space of $x_i$. The second term measures the distribution differences across all modality pairs between the representation space of $x_i$ and held-out samples $h \in H$. Subtracting these terms yields a score that is high when the modalities of $x_i$ are strongly aligned with each other, and weakly aligned with unrelated examples, modeling the model's intended objective.


Finally, we define the \ours of data point $x_i$ as the $\metric$ difference between model $f$ and model $g$:
\begin{equation}
\label{equ:clipmem_multi}
\begin{split}
\text{\textbf{\ours}}(i, H, f) =  \underset{f \sim S}{\metric}(i, H) - \underset{g \sim S ^ {\setminus i}}{\metric}(i, H)\text{.}
\end{split}
\end{equation}

Unlike prior approaches that focus on pairwise similarities between modalities, this resulting metric leverages the entire distribution of representations across all modalities. This provides a principled way to evaluate \textit{global} memorization in multi-modal models, capturing interactions beyond single modality pairs.
\section{Evaluating Multi-Modal Memorization}
We first describe the setup for our experiments and then  analyze memorization in multi-modal contrastive learning using our \ours.

\subsection{Experimental Setup}

\paragraph{\textbf{Models and Datasets.}}\label{sec:model_dataset}
We run our experiments on the following models: OpenCLIP~\cite{cherti2023reproducible}, AudioCLIP~\cite{guzhov2022audioclip}, VideoCLIP~\cite{xu2021videoclip}, and our custom-built \avt (\textbf{A}udio + \textbf{V}ideo + \textbf{T}ext) and \avit (\textbf{A}udio + \textbf{V}ideo + \textbf{I}mage + \textbf{T}ext). The detailed encoder architectures and training datasets used are shown in \Cref{tab:encoder} while other training details and hyperparameters are presented in \Cref{tab:settings} in \Cref{app:ex_exp_setup}. 

\begin{table*}[h]
    \centering         
    \small
    \setlength{\tabcolsep}{4pt}
    
    \resizebox{\textwidth}{!}{\begin{tabular}{cccc}
    \toprule
Model& Modality &Encoder &Training Set\\
    \midrule

 CLIP&Image + Text & ViT + Transformer& COCO~\cite{lin2014microsoft}\\
 VideoCLIP& Video + Text &(CNN + Transformer) + Transformer & MSR-VTT~\cite{xu2016msr}\\
 AudioCLIP& Audio + Image + Text&ESResNeXt + ViT + Transformer & UrbanSound8K~\cite{salamon2014dataset}+ Spectrogram \\
 \midrule
 \avt&Audio + Video + Text &ESResNeXt + (CNN + Transformer) + Transformer&MSR-VTT\\
 \avit&Audio + Video + Image + Text &ESResNeXt + (CNN + Transformer) + ViT + Transformer& MSR-VTT + Frame-image\\
  ImageBind-AVIT& Audio + Video + Image + Text &ESResNeXt + (CNN + Transformer) + ViT + Transformer& MSR-VTT + Frame-image\\
      \bottomrule     
\end{tabular}}
\caption{\textbf{The encoder architecture and datasets used by the models in this paper.} }
\label{tab:encoder}
\end{table*}

\paragraph{\textbf{Dataset Splitting.}}
Following \cite{wang2024memorization}, we divide each training dataset into three
subsets: (1) a candidate set $S_C$, which is used only for training model $f$ and whose memorization we want to measure; (2) an independent set $S_I$, which is used only for training model $g$; and (3) a shared set $S_S$, which is used for training both $f$ and $g$. $S_C$ and $S_I$ have an equal number of training samples to ensure that $f$ and $g$ are trained with the same number of data points.
Detailed splits are provided in \Cref{tab:dataset} in \Cref{app:training}.
We report the average \ours score on $S_C$ as the memorization for model $f$.
Next, we discuss evaluation metrics.

\paragraph{\textbf{Evaluation Metrics.}}
\label{sec:multi_retrieval}
In prior work, retrieval tasks typically refer to settings where one modality is used to retrieve another, which we refer to as the \textit{uni-to-uni} retrieval setting.
To enable the measurement of models' performance on more than two modalities, we introduce a \textbf{multi-to-uni} retrieval task for evaluation. In this task, representations from multiple source modalities are combined to retrieve data points from a target modality. 
Specifically, we compute a retrieval score by summing the pairwise cosine similarities between the target modality representation and \textit{each} source modality representation.
A higher retrieval score shows better semantic alignment between target and input modalities.
A retrieval is considered successful if the ground-truth target sample appears among the top $\textbf{N}$ retrieved samples with the highest retrieval scores; otherwise, it is considered a failure. 
The overall retrieval success rate on the test set is used to evaluate the model’s performance.
In the following experiments, we set $\textbf{N} = 5$ for both uni-to-uni and multi-to-uni retrieval tasks. The final retrieval accuracy is reported using $\text{TOP}@\text{5}$ ($T@5$). 
Note that retrieval tasks are deterministic with no randomness in inference, so we only report the results once instead of an average with standard deviation.
We further rely on \textbf{linear probing} and \textbf{zero-shot} classification tasks to assess model performance under our mitigations. 
For the zero-shot classification, we follow the settings introduced in the CLIP paper~\cite{radford2021learning}, where classification is performed by computing the similarity between label embeddings and the representations of the target modality.



\subsection{Measuring Multi-modal Memorization}
\paragraph{\textbf{Memorization Distribution.}}

\begin{figure*}[t]
    \centering
    \begin{subfigure}[b]{0.45\columnwidth}
        \centering
        \includegraphics[width=\textwidth]{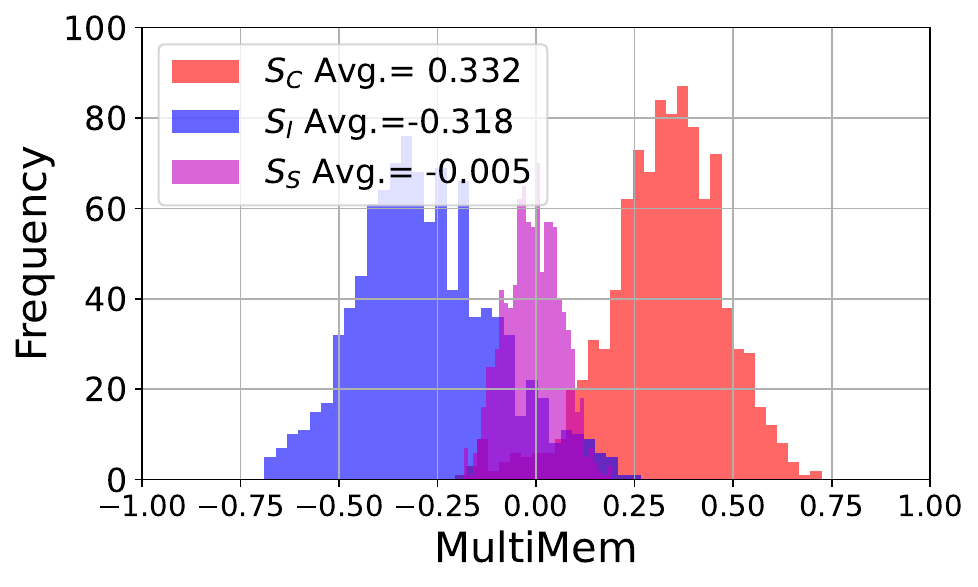}
        \caption{\ours (AIT).}
        \label{fig:multi_vs_clip_0}
    \end{subfigure}
    \begin{subfigure}[b]{0.45\columnwidth}
        \centering
        \includegraphics[width=\textwidth]{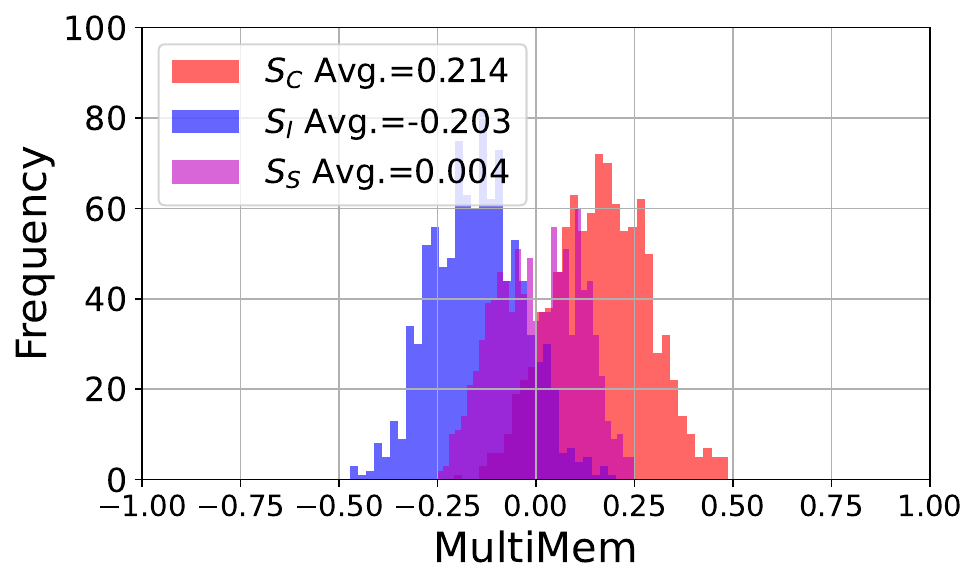}
        \caption{\ours (AI).}
        \label{fig:multi_vs_clip_1}
    \end{subfigure}
    \begin{subfigure}[b]{0.45\columnwidth}
        \centering
        \includegraphics[width=\textwidth]{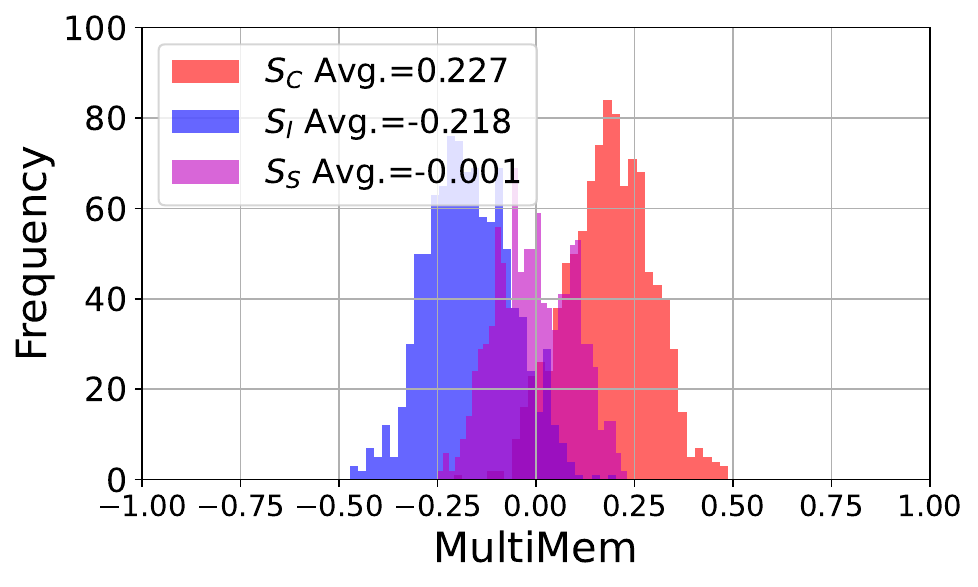}
        \caption{\ours (AT).}
        \label{fig:multi_vs_clip_2}
    \end{subfigure}
    \begin{subfigure}[b]{0.45\columnwidth}
        \centering
        \includegraphics[width=\textwidth]{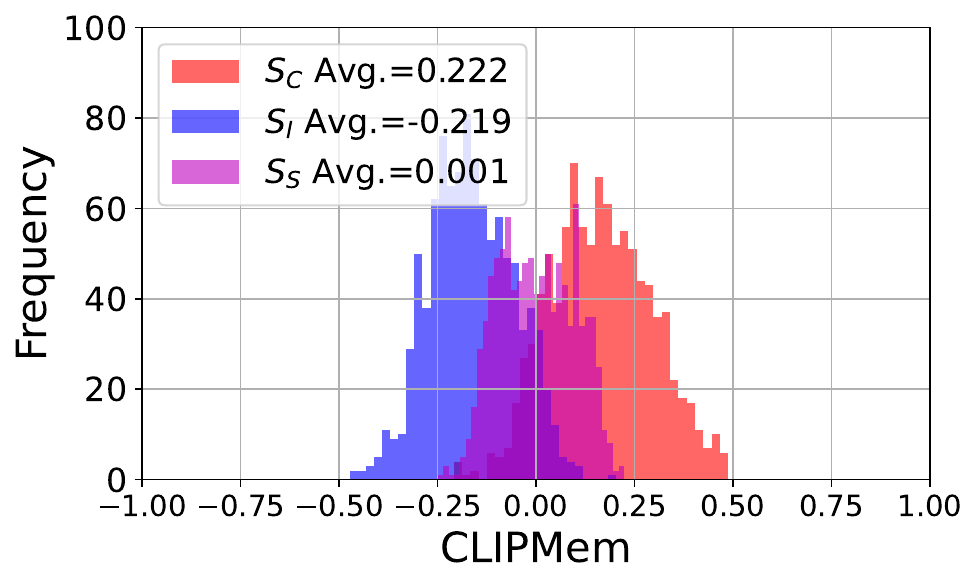}
        \caption{\clip (IT).}
        \label{fig:multi_vs_clip_3}
    \end{subfigure}  
    
    \caption{\textbf{Memorization should be measured on all modalities instead of only on \textit{modality pairs}.} (a) Our \ours scores across all three modalities (AIT: \textbf{A}udio, \textbf{I}mage, and \textbf{T}ext) for AudioCLIP. We quantify pairwise memorization on all modality pairs: (b) \textbf{A}udio-\textbf{I}mage and (c) \textbf{A}udio-\textbf{T}ext (with \ours), and (d) \textbf{I}mage-\textbf{T}ext (with \clip).}
    \label{fig:multi_vs_clip}
\end{figure*}
We study the memorization distribution of training samples on three multi-modal models: AudioCLIP, \avt, and \avit.
For AudioCLIP, we quantify the memorization level with three metrics: (1) tri-modal \ours with all three modalities, (2) the \clip (based on Image-Text pair), and (3) bi-modal \ours restricted to the remaining modality pairs (Audio-Image and Audio-Text). We present our results in \Cref{fig:multi_vs_clip}. 
Overall, \ours in \Cref{fig:multi_vs_clip_0} is able to measure the highest level of memorization on $S_C$ when compared with the baselines on the same AudioCLIP model. 
Additionally, it separates the scores for $S_C$ and $S_S$ significantly better, indicating a more sensitive measurement of memorization.
We observe the same trends for \avt and \avit, as we show in \Cref{fig:all_multi_vs_clip_AVT}, \Cref{fig:all_multi_vs_clip_AVIT}, and \Cref{fig:all_gen_multi_vs_clip_AVIT} in \Cref{app:quantify_avit_gen}.
These findings suggest that in multi-modal models with more than two modalities, a memorization metric that jointly considers all modalities is necessary for a more accurate assessment of memorization.

\begin{figure}[t]
    \centering
    \begin{subfigure}[b]{0.45\columnwidth}
        \centering
        \includegraphics[width=\textwidth]{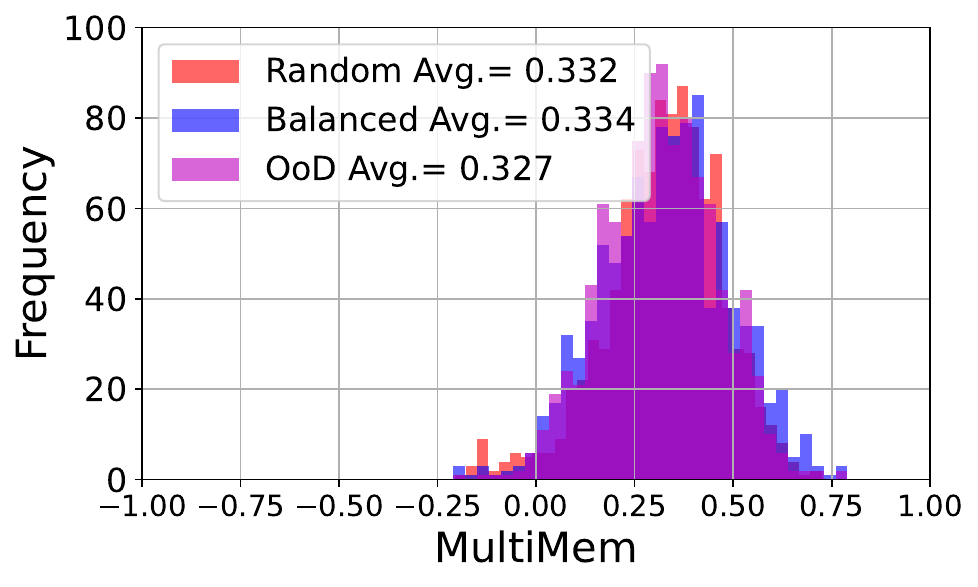}
        \caption{}
        \label{fig:mem_vs_gen_0}
    \end{subfigure}
    \begin{subfigure}[b]{0.45\columnwidth}
        \centering
        \includegraphics[width=\textwidth]{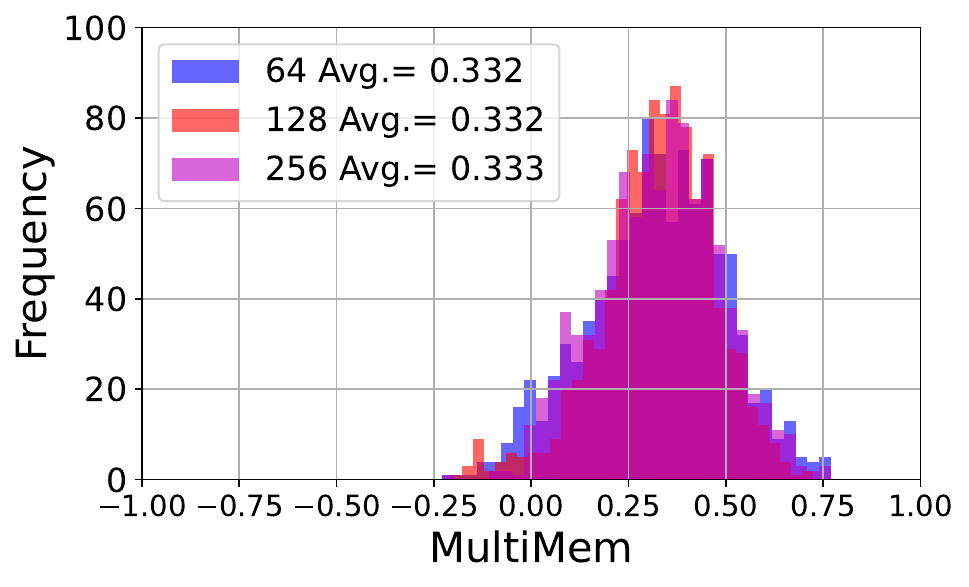}
        \caption{}
        \label{fig:mem_vs_gen_1}
    \end{subfigure}  

    \caption{\textbf{Our MultiMem is robust to hyperparameters $H$.} (a) We varied the composition of $H$ from 128 randomly chosen samples (used in metric) to 128 samples with class balance and 128 OoD samples from the AudioSet dataset \cite{45857}. (b) We varied the number of samples in $H$.}
    \label{fig:robustness}
\end{figure}

\paragraph{\textbf{Robustness to Held-out Set.}}
To examine the sensitivity of our \ours~metric to the choice of held-out set $H$, we evaluate the metric using several selection strategies for $H$: (1) randomly sampled data points, (2) data points chosen to achieve balanced class representation, and (3) samples drawn from an entirely different dataset within the same domain. Furthermore, we investigate the robustness of the metric to the size of $H$ by varying the number of data points included in the set.
The results in \Cref{fig:robustness} show that our MultiMem is robust to the choice of $H$ in both size and composition.

\paragraph{\textbf{Robustness to Dataset Splitting Ratio}}
To validate whether our Multimem is robust to dataset splitting ratio (i.e. $S_S / S_C$ ratio), we added additional experiments on \avt with different dataset splitting ratios on the MSRVTT dataset. We use $S_C = S_I \in \{500, 1500, 2000\}$ in addition to the baseline setting of $S_C = S_I = 1000$ reported in the main paper, while keeping $S_S$ fixed at 6000 throughout. Our results in \Cref{tab:robust_ratio} highlight that reported memorization remains the same over all setups.

\begin{table}[t]
    \centering      
    \small
    \resizebox{\columnwidth}{!}{\begin{tabular}{lcccccc}
    \toprule
    Size of $|S_C|$ &500&1000&1500&2000\\
    \midrule
    Avg. MultiMem&0.405&0.408&0.409&0.408\\
    \bottomrule
\end{tabular}}
\caption{\textbf{Our MultiMem is robust to dataset splitting ratio.}}        
\label{tab:robust_ratio}

\end{table}

\paragraph{\textbf{Highly Memorized Samples.} }
We examine the highly memorized samples in VideoCLIP and \avt to gain deeper insights into the underlying causes of strong memorization. 
For VideoCLIP, we find that, similar to the results reported in the work of \cite{wang2025captured}, videos with misaligned captions and visual content (\ie mis-captioned videos) tend to experience a higher level of memorization.
However, in \avt, we find that highly memorized samples are typically those with semantic misalignment across \textit{multiple} modalities.
These findings highlight the importance of evaluating cross-modal consistency rather than focusing only on text quality. 
Examples of the most memorized samples for VideoCLIP and \avt are shown in \Cref{fig:most_mem} in \Cref{app:top10} and a complete list of the top 10 most memorized samples (including their audio and video content and full captions) is provided in the supplementary material.

\paragraph{\textbf{Memorization and Generalization.}}\label{sec:mem_gen}
\begin{figure*}[t]
    \centering
    \begin{subfigure}[b]{0.35\textwidth}
        \centering
        \includegraphics[width=\textwidth]{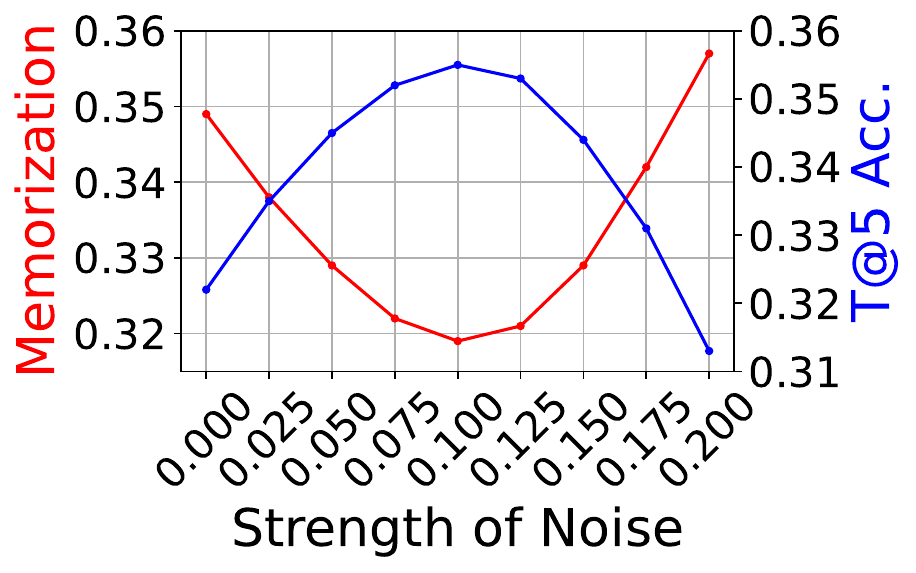}
        \caption{}
        \label{fig:mem_vs_gen_0}
    \end{subfigure}
    \begin{subfigure}[b]{0.35\textwidth}
        \centering
        \includegraphics[width=\textwidth]{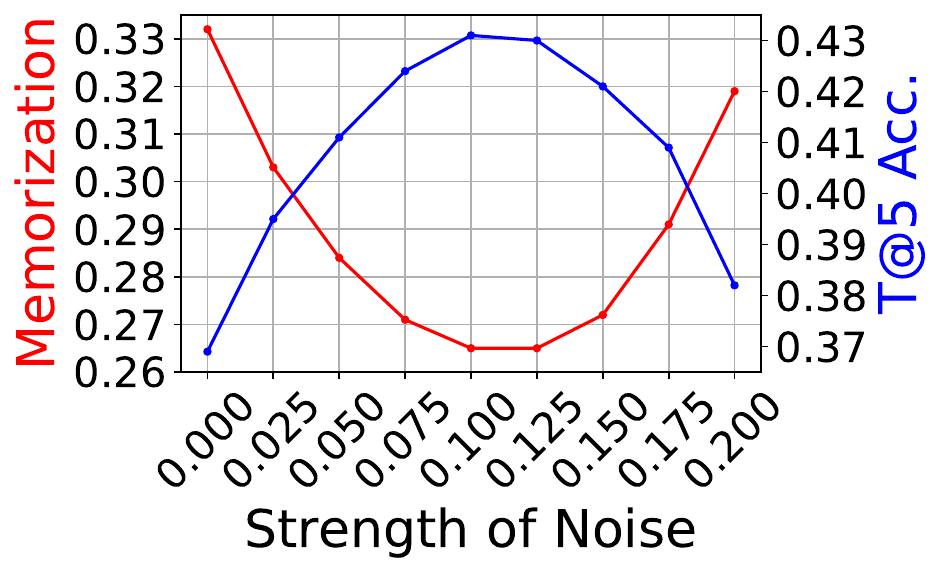}
        \caption{}
        \label{fig:mem_vs_gen_1}
    \end{subfigure}
    
    \begin{subfigure}[b]{0.35\textwidth}
        \centering
        \includegraphics[width=\textwidth]{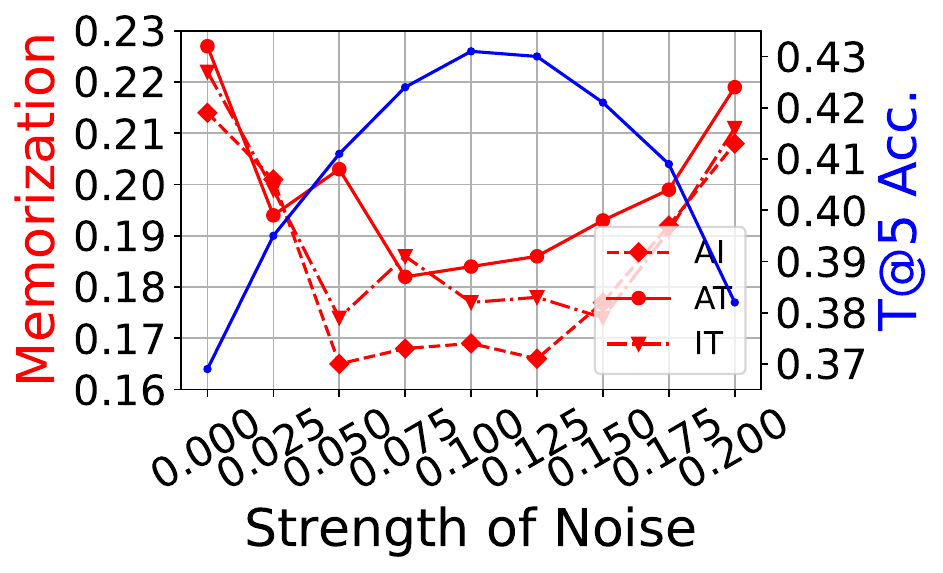}
        \caption{}
        \label{fig:mem_vs_gen_2}
    \end{subfigure}
        \begin{subfigure}[b]{0.35\textwidth}
        \centering
        \includegraphics[width=\textwidth]{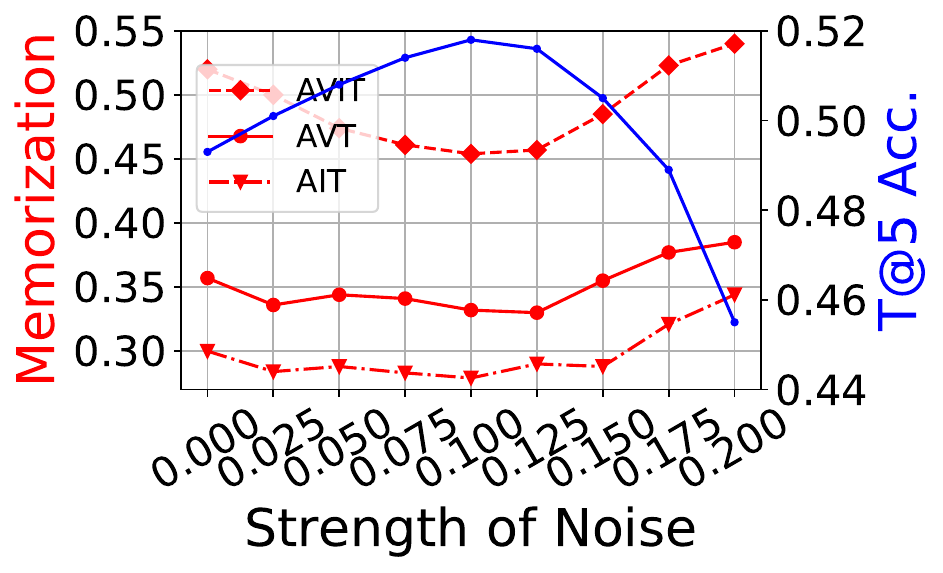}
        \caption{}
        \label{fig:mem_vs_gen_3}
    \end{subfigure}

    \caption{\textbf{Generalization of multi-modal models is negatively correlated with the models' global memorization.} (a) \ours (VT) vs. generalization in VideoCLIP. (b) tri-modal \ours (AIT) vs. generalization in AudioCLIP.
    (c) bi-modal \ours (AI, AT, IT) vs. generalization in AudioCLIP.
    (d) Quad-modal \ours (AVIT) and tri-modal \ours (AVT, AIT) vs. generalization in \avit.}
    \label{fig:mem_vs_gen}
\end{figure*}

To investigate the relationship between generalization and memorization in multi-modal models, we introduce Gaussian noise with different strengths into the representations of all modalities during training as augmentations to control the model's memorization level, as proven successful by previous works \cite{peng2022balanced, yuan2024openvna,wang2025captured}. 
The models' generalization is evaluated by the $T@5$ retrieval task introduced in \Secref{sec:multi_retrieval}.
More specifically, we implement the experiments on VideoCLIP (bi-modal), AudioCLIP (tri-modal) and \avit (quad-modal) by injecting Gaussian noise with mean $\mu=0$ and standard deviations $\sigma =[0.025, 0.050, 0.075, 0.100, 0.125, 0.150, 0.175, 0.200]$.    

In \Cref{fig:mem_vs_gen_0}, when the noise strength is lower than 0.1, we observe that a higher Video-Text \ours results in a worse downstream performance, and vice versa. This aligns with the results reported by \cite{wang2025captured} for bi-modal memorization in CLIP models. 
Moreover, our results in \Cref{fig:mem_vs_gen_1} and \Cref{fig:mem_vs_gen_2} show that AudioCLIP's generalization (trained on three modalities) is negatively correlated with the \textit{tri-modal} \ours. 
In contrast, there is no consistent correlation between generalization and memorization when applying \clip or the bi-modal version of \ours, highlighting again the need to take all modalities into account when assessing multi-modal memorization. We also observe similar trends in \avit for \Cref{fig:mem_vs_gen_3}. 
Together, these results suggest that partial-modal memorization is insufficient to explain how memorization contributes to generalization in multi-modal models and highlight again the benefits of a truly multi-modal metric. 

Finally, when the noise strength exceeds 0.1, the performance of both AudioCLIP and VideoCLIP begins to decline, while global memorization increases. This occurs because excessive noise destroy the semantic consistency between modalities. 
As a result, learning across modalities becomes more difficult, leading to increased global memorization and weakening the model’s generalization.

\paragraph{\textbf{Impact of Augmentations.} }\label{sce：aug}
We further study how applying augmentations (injecting noise) to different modality configurations (\eg single modality, a subset of all modalities, and all modalities) during training influences the generalization of multi-modal models. 
We adopt SSLMem from \cite{wang2024memorization} to measure memorization for single modality and use \ours for multi-modal memorization. 
The results in \Cref{tab:augmentation_adclip} on AudioCLIP suggest that applying augmentations to a single modality slightly increases the SSLMem of all modalities and enhances the model's global generalization. 
However, it also leads to increased memorization in modality pairs that do not involve the augmented modality, which in turn degrades the performance of retrieval tasks based on that modality pair (as indicated by the underlined values in the tables). 
Moreover, applying augmentations to a subset of the modalities reduces both global and pairwise memorization, resulting in improved performance over all retrieval tasks. 
Notably, applying augmentations to all modalities always yields the best performance and the lowest memorization (as reflected by the bolded values in the tables), which provides a foundation for the \ours-based memorization mitigation strategies that we introduce in \Secref{sec:mem_mitigate}.

\begin{table*}[t]
    \centering      
    \small
    \resizebox{\textwidth}{!}{\begin{tabular}{cccccccccccc}
    \toprule

Augmented modality&
\makecell{IT\\ \ours} &\makecell{AT\\ \ours} & 
\makecell{AI\\ \ours} &\makecell{AIT\\ \ours} &\makecell{A\\SSLMem}&\makecell{I\\SSLMem}& \makecell{T\\SSLMem} & \makecell{T@5(\%)\\I-T} &\makecell{T@5(\%)\\A-T}&\makecell{T@5(\%)\\A-I}&\makecell{T@5(\%)\\AI-T}\\
    \midrule
None&0.222&0.227&0.214&0.332&0.188&0.191&0.210&33.1&30.8&26.5&36.9\\
\midrule
Audio&\underline{0.235}&0.201&0.204&0.320&0.199&0.196&0.214&\underline{32.2}&31.3&27.1&38.2\\
Image&0.201&\underline{0.235}&0.188&0.312&0.194&0.210&0.217&33.8&\underline{30.0}&28.3&39.0\\
Text&0.181&0.191&\underline{0.216}&0.294&0.196&0.201&0.235&34.4&32.1&\underline{26.3}&39.8\\
\midrule
Audio + Image&0.200&0.199&0.199&0.299&0.207&0.219&0.233&33.9&31.8&27.5&39.4\\
Audio + Text&0.187&0.187&0.192&0.283&0.228&0.211&0.244&34.9&32.6&28.0&40.7\\
Image + Text&0.180&0.189&0.174&0.271&0.211&0.240&0.251&35.4&32.3&29.1&42.1\\
\midrule
Audio + Image + Text&\textbf{0.177}&\textbf{0.184}&\textbf{0.169}&\textbf{0.264}&0.236&0.262&0.269&\textbf{35.7}&\textbf{33.0}&\textbf{29.5}&\textbf{43.1}\\
\bottomrule     
\end{tabular}}
\caption{\textbf{Adding augmentations to different modalities in AudioCLIP.} We add random Gaussian noise with $\mu=0$, $\sigma=0.1$ to the representations as augmentations. \textbf{A}: Audio, \textbf{I}: Image, and \textbf{T}: Text.}        
\label{tab:augmentation_adclip}
\end{table*}

\paragraph{\textbf{Memorization with ImageBind.}}
We train an additional AVIT model using the ImageBind-loss~\cite{girdhar2023imagebind}. Unlike our setup, which performs pairwise alignment of \emph{all} modalities, the ImageBind-loss takes the image representation from a pretrained image encoder as a reference and aligns all other modalities exclusively to this representation.
During training, the image encoder remains frozen. 
Our results show that training with the ImageBind loss leads to increased global memorization (0.571 versus 0.488 achieved with our original AVIT) and reduced performance in retrieval task (36.5\% versus 48.2\% our AVIT model). 
These results suggest that training with a pairwise alignment across \textit{all} modalities is beneficial as it reduces memorization and improves generalization.
In \Cref{app:ex-exp}, we present further details and an additional results on how the integration of more modalities during training improves generalization.

\paragraph{\textbf{Balancing Per-Modality Memorization.}}\label{sec:balance}

Given the differences in how each modality contributes to the overall memorization (see for example \Cref{tab:augmentation_adclip} for AudioCLIP), we further show that enforcing a balanced memorization distribution among modalities during training leads to a lower overall memorization and improved model performance.  We apply $\ell_1$-normalized weights of $[0.332, 0.324, 0.344]$, which are negatively correlated to their memorization contribution $[0.222, 0.227, 0.214]$, to the three modality pairs (I-T, A–T, A–I) for AudioCLIP training. The new trained model achieves a lower MultiMem of 0.290 (compare to 0.332 of baseline model), a 3.2\% improvement in retrieval task performance and a more balanced bi-modal MultiMem level of $[0.192, 0.195, 0.193]$ for three modality pairs (I-T, A–T, A–I).
We present the full setup in \Cref{app:ex-exp}.

\paragraph{\textbf{Memorization Behavior in Multi-modal, SSL and SL.}}
We apply \unit~\cite{wang2024localizing} to analyze where inside the models memorization happens. We compare AudioCLIP (tri-modal), and compare it with CLIP (bi-modal), SL, and SSL models (uni-modal).
All these models are trained with UrbanSound8K dataset + spectrogram images.
A higher average \unit at a given layer indicates a greater contribution of that layer to the model’s global memorization.

\begin{figure}[t]
\centering
    \includegraphics[width=0.35\textwidth]{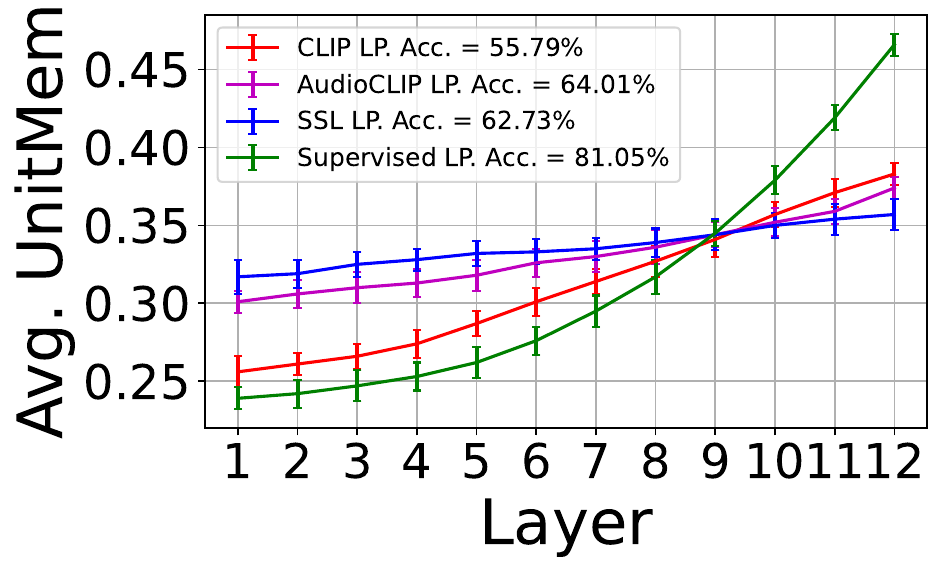}
    \caption{\textbf{\unit: AudioCLIP is more aligned with SSL.} }
    \label{fig:image_unitmem}
\end{figure}

We observe that, compared to CLIP, the vision encoder in AudioCLIP aligns more with SSL, rather than falling between SSL and SL as CLIP.
This suggests that the introduction of the audio modality partially reduces the dominance of the text modality (\eg with labels or captions) during training.
Thereby, the memorization behaviors of the vision encoder shifts from being label-driven (in SL) or caption-driven (in CLIP), to pattern-driven as in SSL~\cite{wang2024memorization}. 
We hypothesize that this is because the learning signals in multi-modal models no longer come from a single supervision source (\eg labels or captions). Instead, the model focuses more on the shared or similar semantic patterns \textit{across modalities} such as audio, vision, and text. Therefore, the pattern underlying this shift is the \textit{strong semantic interdependence} across modalities in multi-modal models.


\section{Our Mitigation Strategies}\label{sec:mem_mitigate}
Our findings from the previous section highlight that mitigating cross-modal memorization over all modalities can improve the generalization of multi-modal models. Based on this motivation, we design two methods built on \ours to mitigate high memorization, either during or post training.

\begin{figure*}[!t]
    \centering
    \begin{subfigure}[b]{0.64\columnwidth}
        \centering
        \includegraphics[width=\textwidth]{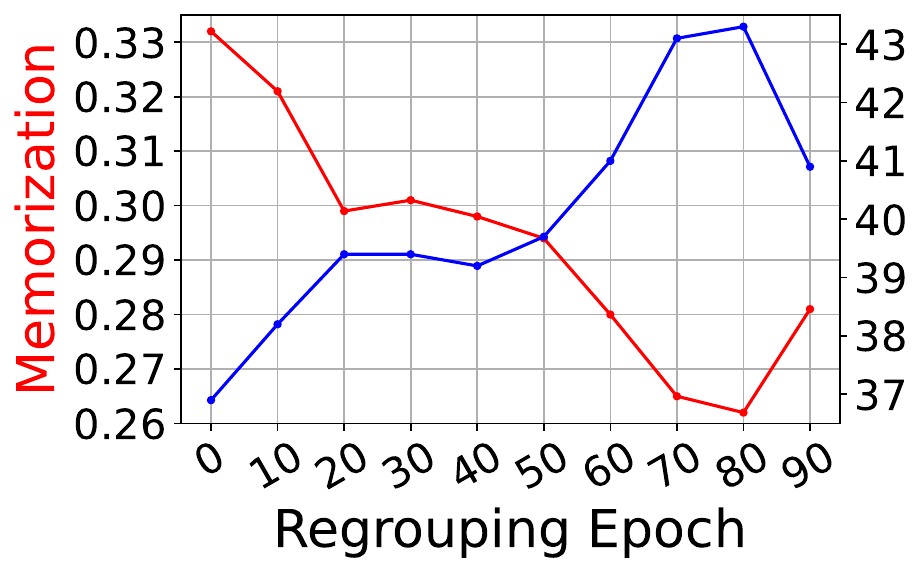}
        \caption{Retrieval (T@5\%)}
        \label{fig:regrouping1}
    \end{subfigure}
    \centering
    \begin{subfigure}[b]{0.64\columnwidth}
        \centering
        \includegraphics[width=\textwidth]{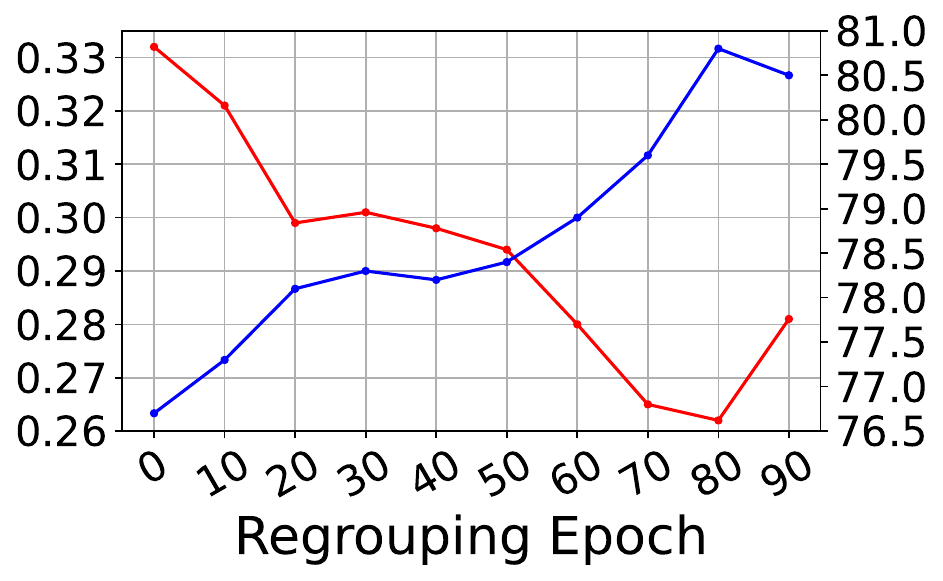}
        \caption{Linear Probing (\%)}
        \label{fig:regrouping2}
    \end{subfigure}
    \centering
    \begin{subfigure}[b]{0.64\columnwidth}
        \centering
        \includegraphics[width=\textwidth]{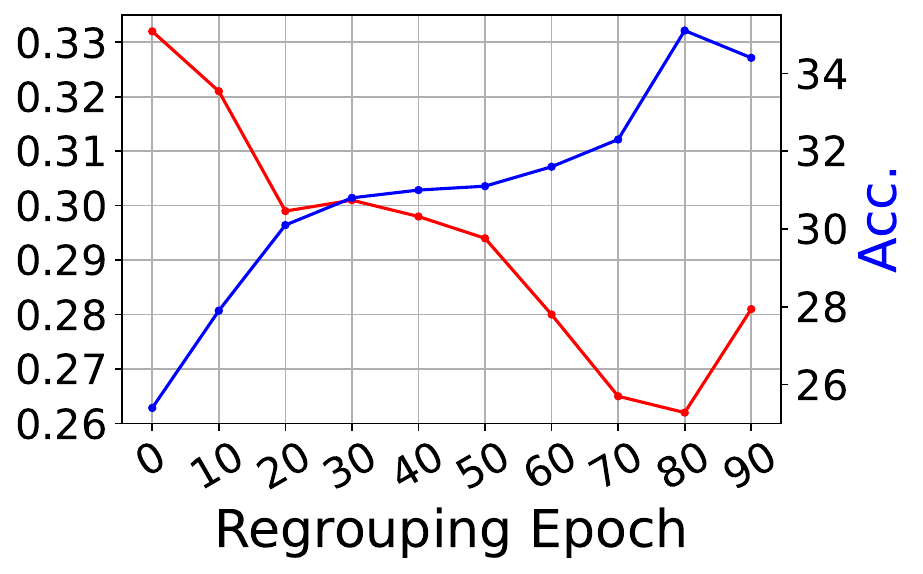}
        \caption{Zero-shot (\%)}
        \label{fig:regrouping3}
    \end{subfigure}
   \\
    \begin{subfigure}[b]{0.64\columnwidth}
        \centering
        \includegraphics[width=\textwidth]{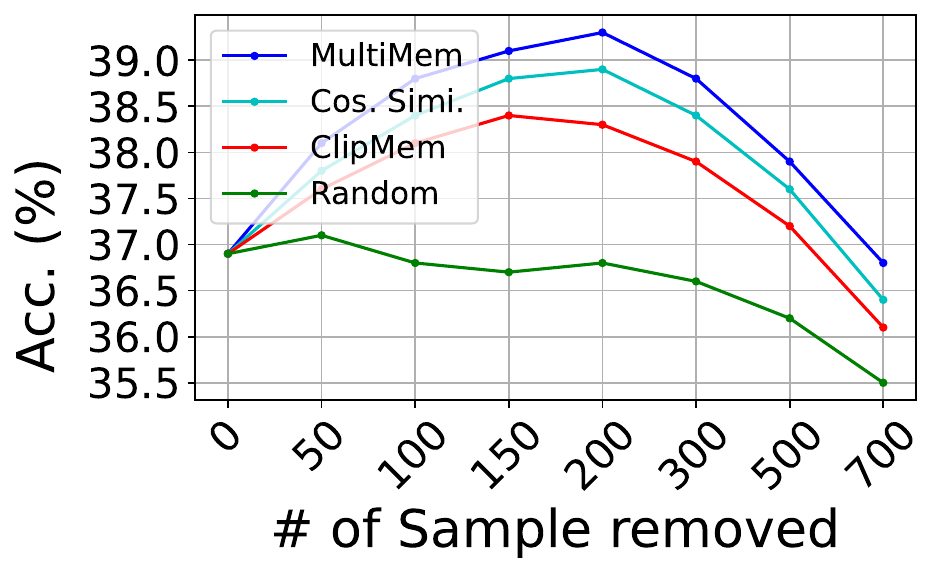}
        \caption{Retrieval (T@5\%)}
        \label{fig:removing1}
    \end{subfigure}
     \begin{subfigure}[b]{0.64\columnwidth}
        \centering
        \includegraphics[width=\textwidth]{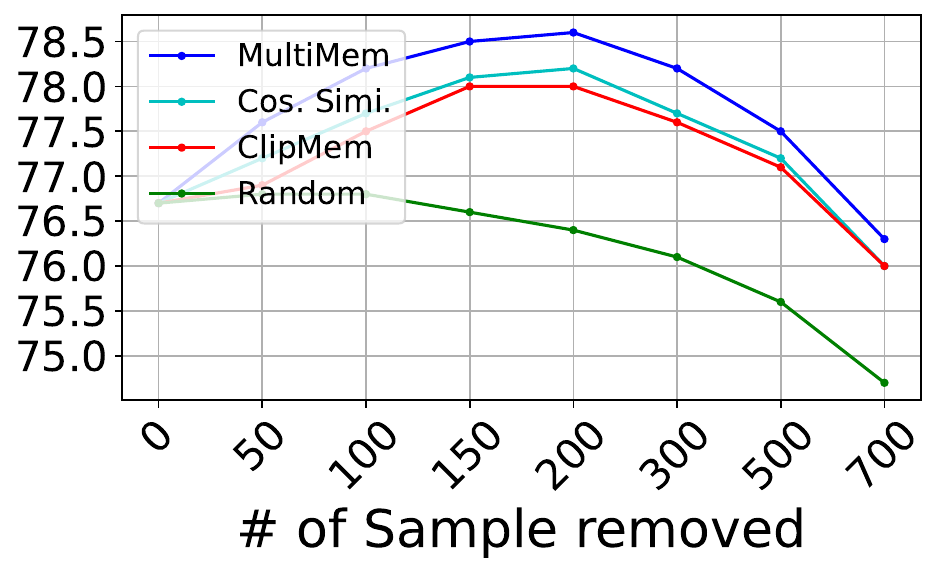}
        \caption{Linear Probing (\%)}
        \label{fig:removing2}
    \end{subfigure}  
     \begin{subfigure}[b]{0.64\columnwidth}
        \centering
        \includegraphics[width=\textwidth]{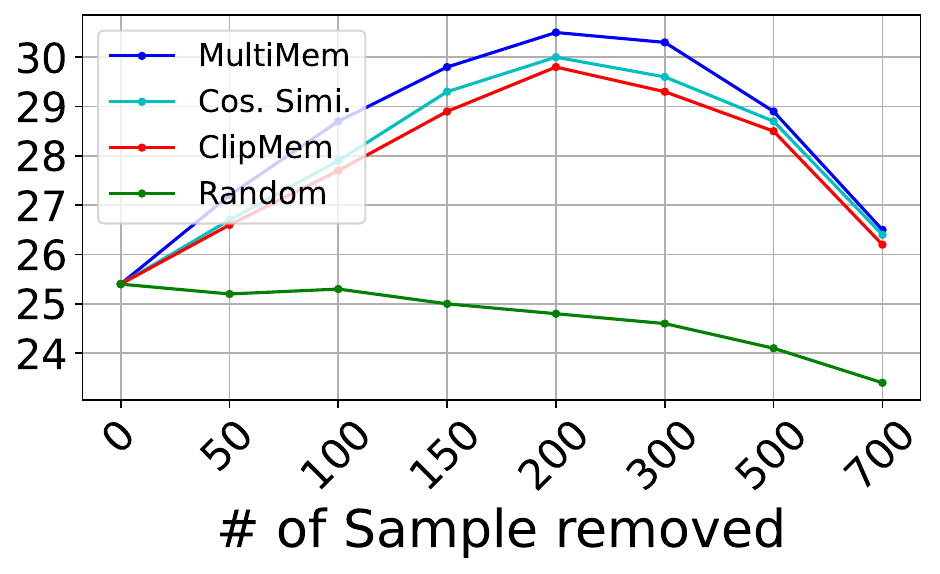}
        \caption{Zero-shot (\%)}
        \label{fig:removing3}
    \end{subfigure}  
    \caption{
    \textbf{Mitigation Strategies.} The first row shows the impact of in-training memorization mitigation at different training epochs, evaluated on (a) retrieval, (b) linear probing, and (c) zero-shot classification tasks. The second row presents the effect of post-training mitigation with increasing number of most memorized samples removed, analogously for (d) retrieval, (e) linear probing, and (f) zero-shot classification tasks.
    }
    \label{fig:mitigation}
\end{figure*}

\paragraph{\textbf{In-Training Mitigation.}}
Instead of applying noise to the entire dataset, which harms overall performance and incurs additional computational overhead, 
we selectively add Gaussian noise ($\mu=0$, $\sigma=0.1$) as augmentation only to the representations of the most memorized samples during training to mitigate memorization. 
Specifically, at every 10-epoch interval (\ie epochs 10, 20, ..., 90), we use \ours to measure memorization for all training samples and select the top 5\% most memorized samples. 
The 5\% ratio is selected based on our experiments in \Cref{fig:augment_ratio} in \Cref{app:augment_ratio}, which show that this ratio yields the highest model performance and the lowest memorization level.
We then aggregate these samples into new mini-batches and apply noise-based augmentations to their representations during the following training steps while keeping training unaltered for all other mini-batches. 

In our experiment, we train AudioCLIP on the UrbanSound8K + Spectrogram dataset for 100 epochs with our mitigation in place and report performance on the retrieval task in \Cref{fig:regrouping1}, the classification task on UrbanSound8K in \Cref{fig:regrouping2}, and the zero-shot classification task on AudioSet~\cite{45857} in \Cref{fig:regrouping3}.

The results yield two main findings. First, applying our mitigation strategy at any training stage effectively reduces global memorization and improves overall performance across all three tasks.
Second, model performance in all three tasks follows a trend of initial improvement, followed by a plateau, then further improvement, with a slight decline observed at epoch 90.
We attribute this pattern to the timing of memorization mitigation. Applying mitigation too early in training may fail to accurately capture the most memorized samples, as the model is far from convergence. As a result, highly memorized examples that are not identified in the early stages may continue to increase the model’s memorization in later epochs, thereby harming generalization. Applying memorization mitigation too close to the end of training may lead to insufficient decoupling of modality correlation for highly memorized samples. This is the reason why the performance drop is observed at epoch 90 compared to epoch 80. 
These results indicate that applying noise-based augmentation near the end of training (\eg at 80\%) to a selected ratio of most memorized samples according to \ours can effectively reduce memorization and enhance generalization.
In \Cref{app:mit_mem_intrain}, we provide further insights into this strategy. (1) We compare the mitigation effects between using noise-based augmentation and gradient clipping (which is widely used in Differential Privacy area) for most memorized samples.   (2) We compare it with an alternative approach that directly removes a fixed proportion of the most memorized samples during training. The results show that our strategy yields better performance improvements. (1) We also examine the effect of repeatedly applying this strategy at multiple stages of training. The findings indicate that repeated application leads to additional gains compared to single use. However, the improvement becomes marginal when applied more than twice.


\paragraph{\textbf{Post-Training Mitigation.}}

\begin{figure*}[t]
    \centering
    \begin{subfigure}[b]{0.35\textwidth}
        \centering
        \includegraphics[width=\textwidth]{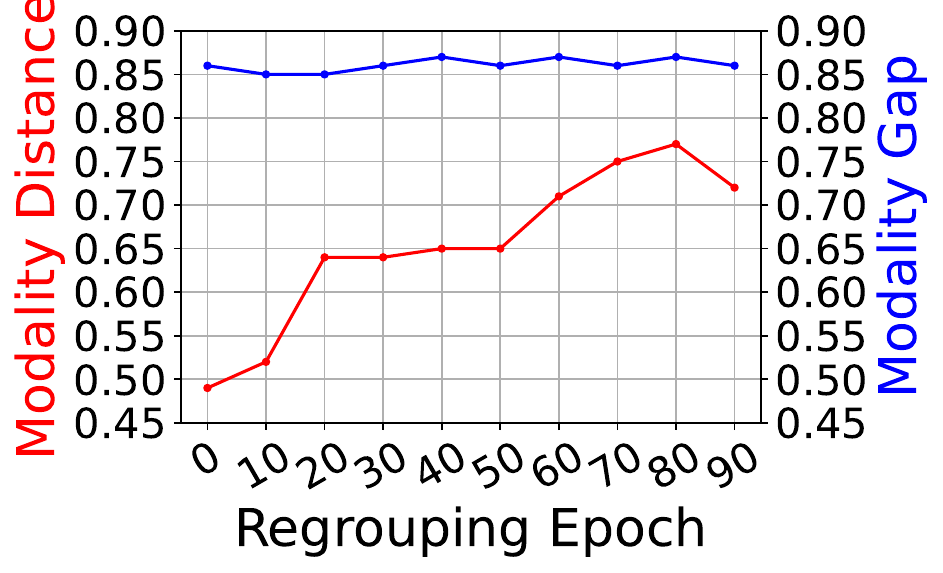}
        \caption{In-training Mitigation}
        \label{fig:gap0}
    \end{subfigure}
    \begin{subfigure}[b]{0.35\textwidth}
        \centering
        \includegraphics[width=\textwidth]{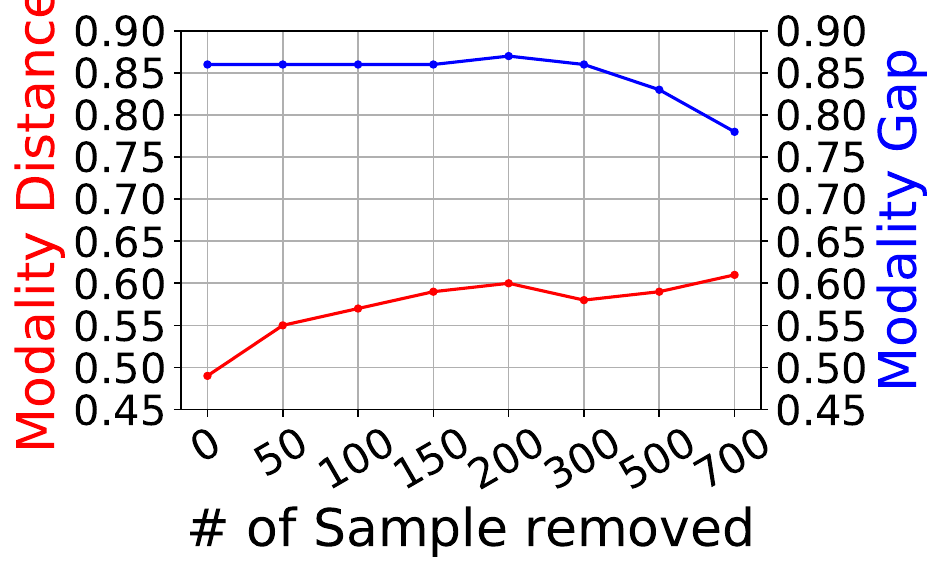}
        \caption{Post-training Mitigation}
        \label{fig:gap1}
    \end{subfigure}  

    \caption{\textbf{Average modality representation distance for 1\% most memorized samples versus averagae modality gap for all training samples before and after mitigation} }
    \label{fig:gap}
\end{figure*}

Post-training, we first train the model and then use \ours to identify the most memorized samples. Then, we remove these samples from training and fine-tune the model for some additional steps on the remaining data.
In our experiments, we train an AudioCLIP model on the UrbanSound8K dataset for 100 epochs, then, we use \textit{\ours} to identify the most memorized samples and remove the top [50, 100, 150, 200, 300, 500, 700] of them. 
Finally, we fine-tune the model with the remaining dataset for an additional 25 epochs.
For comparison, we additionally employ three alternative strategies for selecting samples to remove as baselines: using \textit{\clip}, computing the sum of \textit{cosine similarities} across modality pairs, and \textit{randomly} selecting samples.

We show the results in \Cref{fig:mitigation} (lower row) for retrieval, classification on UrbanSound8K, and zero-shot tasks accuracy on AudioSet. 
We observe that regardless of the method used to mitigate the model's memorization (apart from \textit{random}), the model's performance improves to varying degrees when fewer than 500 samples are removed on all three tasks. Among them, \textit{\ours} yields the best performance improvement, followed by the \textit{cosine similarity} and \textit{\clip}. In contrast, the baseline of removing \textit{random} samples does not show a clear trend of performance improvement. This highlights the effectiveness of \ours in post-training memorization mitigation and improving generalization.

\paragraph{Regularization vs Memorization Mitigation.}
Prior work has shown that Gaussian noise may act as a regularizer that independently improves performance rather than genuinely mitigating memorization~\cite{feldman2020does,feldman2020neural,yuan2024openvna}. To verify that the observed performance improvements and memorization mitigation are actually from our proposed mitigation strategy rather than the generic regularization effect of noise, we conducted new experiments where we add \textit{the same amount of noise} to different types of samples. 1) most memorized samples and least memorized samples identified by MultiMem, 2) samples with highest gradients or loss, 3) randomly selected samples. We report the performance gains on retrieval, linear probing, and zero-shot tasks, and MultiMem. 
If the improvements were only due to generic regularization rather than memorization mitigation, we would observe similar results over all setups. However, the results in the~\Cref{tab:regulation} show discrepancies: 1) adding noise to random samples does not change performance significantly. 2) Adding noise to least memorized samples slightly degrades performance. 3) Our \textbf{MultiMem approach has much higher performance gains compared to loss-based and gradient-based methods and also achieves the best memorization mitigation.} This shows that our reported improvements indeed stem from memorization removal and not general regularization.

\begin{table*}[t]
    \centering
\resizebox{\textwidth}{!}{\begin{tabular}{l|cccc|cccc}
    \toprule
    & \multicolumn{4}{c|}{\textbf{Post-training}} & \multicolumn{4}{c}{\textbf{In-training}} \\
    \cmidrule(lr){2-5} \cmidrule(lr){6-9}
    \textbf{Mitigation} & Retrieval$\uparrow$ & Linear Prob.$\uparrow$ & Zero-Shot$\uparrow$ &Mem.$\downarrow$& Retrieval$\uparrow$ & Linear Prob.$\uparrow$ & Zero-Shot$\uparrow$ &Mem.$\downarrow$\\
    \midrule
    None & 36.9\% & 76.7\% & 25.4\% & 0.332&36.9\% & 76.7\% & 25.4\% &0.332\\
    \cdashline{1-9}
    Random &37.1\%&76.8\%&25.5\%&0.330  &37.2\%&76.9\%&25.5\%&0.329\\
    MultiMem (least)&36.6\%&76.6\%&25.2\%&0.336 &36.1\%&76.1\%&25.0\%&0.338\\
    \cdashline{1-9}
    Gradient & 38.1\% & 77.8\% & 29.4\%& 0.307& 40.4\% & 78.8\% & 31.1\%&0.281\\
    Loss & 38.4\% & 77.9\% & 29.6\% & 0.308&39.9\% & 78.9\% & 30.9\% &0.280\\
    \textbf{MultiMem (most)} & \textbf{39.3\%} & \textbf{78.6\%} & \textbf{30.5\%} &\textbf{0.282}& \textbf{43.3\%} & \textbf{80.8\%} & \textbf{35.1\%} &\textbf{0.262}\\
    \bottomrule     
\end{tabular}}
\caption{\textbf{Regulation verification by comparing MultiMem with other mitigation strategy.} }
\label{tab:regulation}
\end{table*}

\begin{table*}[t]
    \centering
\resizebox{\textwidth}{!}{\begin{tabular}{l|cccc|cccc}
    \toprule
    & \multicolumn{4}{c|}{\textbf{Post-training}} & \multicolumn{4}{c}{\textbf{In-training}} \\
    \cmidrule(lr){2-5} \cmidrule(lr){6-9}
    \textbf{Mitigation} & Retrieval$\uparrow$ & Linear Prob.$\uparrow$ & Zero-Shot$\uparrow$ &Mem.$\downarrow$& Retrieval$\uparrow$ & Linear Prob.$\uparrow$ & Zero-Shot$\uparrow$ &Mem.$\downarrow$\\
    \midrule
    None & 36.9\% & 76.7\% & 25.4\% & 0.332&36.9\% & 76.7\% & 25.4\% &0.332\\
    \cdashline{1-9}
    \textbf{Noise} & \textbf{39.3\%} & \textbf{78.6\%} & \textbf{30.5\%} &\textbf{0.282}& \textbf{43.3\%} & \textbf{80.8\%} & \textbf{35.1\%} &\textbf{0.262}\\
    Augmentation&38.8\%&78.4\%&30.0\%&0.288&42.5\%&80.1\%&34.0\%&0.269\\
    \bottomrule     
\end{tabular}}
\caption{\textbf{Modality-specific augmentation vs. noise injection.} }
\label{tab:aug_noise}
\end{table*}

\paragraph{\textbf{Insights into Mitigation.}}
We further compared the average modality distance between the representations of each modality for the most memorized samples, as well as the average modality gap~\cite{liang2022mind} across all samples in the training set, before and after mitigation.
Results in \Cref{fig:gap} show that: (1) The model struggles to align semantic misaligned samples through generalization, thereby memorizes those datapoints so that results in an extremely low average sample modality distance. This shows that semantic misalignment is the key driver of memorization in multi-modal models. (2) Both of our mitigation methods increase the average modality distance of the most memorized samples thereby reducing the memorization level of the model. However, in-training mitigation performs better than post-training mitigation, and therefore yields a greater improvement in the generalization of the model. At the same time, in-training mitigation preserves the original average modality gap better. This is consistent with the conclusion in \cite{liang2022mind}, that preserving the original modality gap tends to yield better model performance. These results show that our mitigation methods address the fundamental cause of memorization in multi-modal models, \ie controlling the average modality distance for semantic misaligned samples. 

\paragraph{Impact of Noise Injection vs. Augmentation.}
Prior work~\cite{peng2022balanced, yuan2024openvna,wang2025captured} has shown that Gaussian noise can indeed serve as a generic augmentation strategy across modalities. This, however, raises the concern of whether such noise genuinely achieves an effect comparable to that of modality-specific augmentations, and whether it yields a similar level of memorization mitigation. To address this concern, we conduct an additional control experiment on AudioCLIP. Specifically, we replace noise injection with modality-specific augmentations: Gaussian blur (sigma=(0.1, 2.0)) combined with random color jitter (transforms.ColorJitter(0.4, 0.4, 0.4, 0.1)) for the image modality, synonym substitution (p=0.2) for captions, and time shifting (+10\%) for audio. As shown in \Cref{tab:aug_noise}, mitigation guided by these modality-specific augmentations remains effective, achieving performance comparable to that of noise injection. This indicates that the mitigation effect arises from the underlying mechanism itself rather than from the generic regularization properties of noise.

\section{Conclusions}
We propose \ours, a novel metric for measuring and characterizing memorization in multi-modal contrastive learning with arbitrary modality configurations. 
Our results show clear differences compared to commonly studied bi-modal models like CLIP.
We find that training with multiple modalities not only mitigates the single-modality dominance in global memorization observed in bi-modal models, but also makes the contrastive models' memorization behavior more similar to that of self-supervised learning.
Specifically, text modality, which has been viewed in previous work~\cite{wang2025captured} as similar to labels, is no longer the only reason for high memorization when it does not align with other modalities. 
Inconsistency in semantic information between multiple modality pairs is instead the leading cause of high memorization. 
Finally, we show that our proposed \ours metric can be used to inform mitigations against memorization, either during training or after training, that improve generalization more efficiently than other metrics. 

\section*{Acknowledgments}
This research was funded by the Deutsche Forschungsgemeinschaft (DFG, German Research Foundation), Project number 550224287. Franziska Boenisch received funding from the European Research Council (ERC) under the European Union’s Horizon Europe research and innovation programme (grant agreement No 101220235). We would like to acknowledge our sponsors, who support our research with financial and in-kind contributions: OpenAI and G-Research. We also thank members of the SprintML group for their
feedback. 




%
%

{
    \small
    \bibliographystyle{ieeenat_fullname}
    \bibliography{main}
}

\clearpage
\appendix
\section{Appendix}\label{app:appendix}

\subsection{Hardware Usage \& Calculation Overhead}\label{app:hardware}
Two devices are used for all experiments mentioned in this paper: a cloud
server with four A100 GPUs and a local workstation with Intel 13700k CPU, Nvidia 4090 graphics card, a total of 10Tb storage space and 64GB of RAM.

Applying In-training mitigation strategy once during training will bring 0.70\% (2.6 min out of 372 min) overhead in total time usage, without requiring any additional memory. Applying Post-training mitigation strategy will bring 0.55\% (2.1 min out of 372 min) overhead in total time usage for memorization measurement, without requiring any additional memory.

Measuring \ours for all samples in candidate set $S_C$ averagely brings 0.48\% overhead in total time usage, without requiring any additional memory. Since the computational complexities of our \ours  on both the number of modalities and the number of training samples are $\text{O}(n^2)$, which matching the complexity of multi-modal contrastive learning. Therefore, the relative overhead remains essentially unchanged as modalities or data scale up.

\subsection{Extended Experiment Setup}\label{app:ex_exp_setup}

\paragraph{\textbf{Glossary.}} For the reader's convenience, we provide a glossary with all important notation used in the main paper in \Cref{tab:notation}.

\begin{table*}[t]
    \centering            
    \small
    
    \begin{tabular}{cc}
    \toprule
Symbol & Explanation\\

 \midrule
A & Audio modality\\
V & Video modality\\
I & Image modality\\
T & Text modality\\
AVT-CLIP & Variant of CLIP with Audio + Video + Text modalities\\
AVIT-CLIP & Variant of CLIP with Audio + Video + Image + Text modalities\\
H & Held-out test set used for \ours\\
Aug & Augmentation set for data points\\
      \bottomrule     
\end{tabular}

\caption{\textbf{We provide an overview on the important notation in main paper.} } 
\label{tab:notation}
\end{table*}

\paragraph{\textbf{Dataset Split.}}\label{app:training}
\Cref{tab:dataset} provides a detailed summary of the splitting $S_C$, $S_I$, and $S_S$ discussed in main paper.
\begin{table}[t]
    \centering            

    \resizebox{0.5\textwidth}{!}{\begin{tabular}{ccccc}
    \toprule
Dataset& Total &$S_S$& $S_I$ & $S_C$\\
 \midrule
   COCO&123287&65000&5000&5000\\
   MSR-VTT&10000&7000&1000&1000\\
   UrbanSound8K&8732&6000&1000&1000\\
      \bottomrule     
\end{tabular}}

\caption{\textbf{Detailed dataset split used in this paper.} } 
\label{tab:dataset}
\end{table}

\paragraph{\textbf{Training Setup.}}\label{app:training}
\Cref{tab:settings} provide a detailed summary of the training configurations for all models used in our experiments. As stated in the main paper, all settings for existing models follow the default configurations of their respective original implementations.  
\begin{table*}[t]
    \centering
    \small 
  
\resizebox{\textwidth}{!}{\begin{tabular}{cccccccc}
\toprule
& SL-ViT & DINO & CLIP &VideoCLIP &AudioCLIP &\avt &\avit \\ 
                       \midrule
Training Epochs &100&300&100&100&100&100&200\\
Warm-up Epochs   &5&20&10&10&10&10 &20     \\
Batch Size       &1024&1024&512&256&64&64&64   \\
Optimizer          &AdamW&AdamW&AdamW&Adam&Adam&Adam&Adam\\
Learning rate     &1e-3&2e-3&1e-3&1e-4&1e-4&1e-4&1e-4\\
Learning rate Schedule &Cos. Decay&Cos. Decay&Cos. Decay&Cos. Decay&Cos. Decay&Cos. Decay&Cos. Decay\\ 
\bottomrule 
\end{tabular}}
    \caption{\textbf{Training Setup and Hyperparameter for all models used in this work.} }     
          \label{tab:settings}
\end{table*}

\paragraph{\textbf{\avt and \avit.}}\label{app:avtclip}
For \avt, the video and text modalities are first pre-trained for 20 epochs, after which the audio modality is introduced. The model is then trained jointly on all three modalities for an additional 100 epochs. For \avit, the image and text modalities are first pre-trained for 20 epochs, after which the video and audio modalities are introduced and trained jointly on all four modalities for an additional 200 epochs. 

\paragraph{\textbf{Experimental Setup for \unit.}}\label{app:sl_ssl_multi}
In this experiment, we use the original UrbanSound8K dataset as the source of both audio inputs and text labels. Following the evaluation setup of AudioCLIP, we also use spectrograms of the audio signals as image inputs. In the SL setting, we employ a ViT encoder followed by a two-layer MLP classifier, trained with spectrogram images as inputs and text labels as supervision. For SSL settings, we employ the ViT-based DINO framework and train the ViT encoder on only spectrogram images. Both the image augmentation sets and hyperparameters follow the default configurations of their respective original implementations.

\subsection{Extended Experiment Results}\label{app:ex-res}

\begin{figure*}[t]
    \centering
    \begin{subfigure}[b]{0.45\textwidth}
        \centering
        \includegraphics[width=\textwidth]{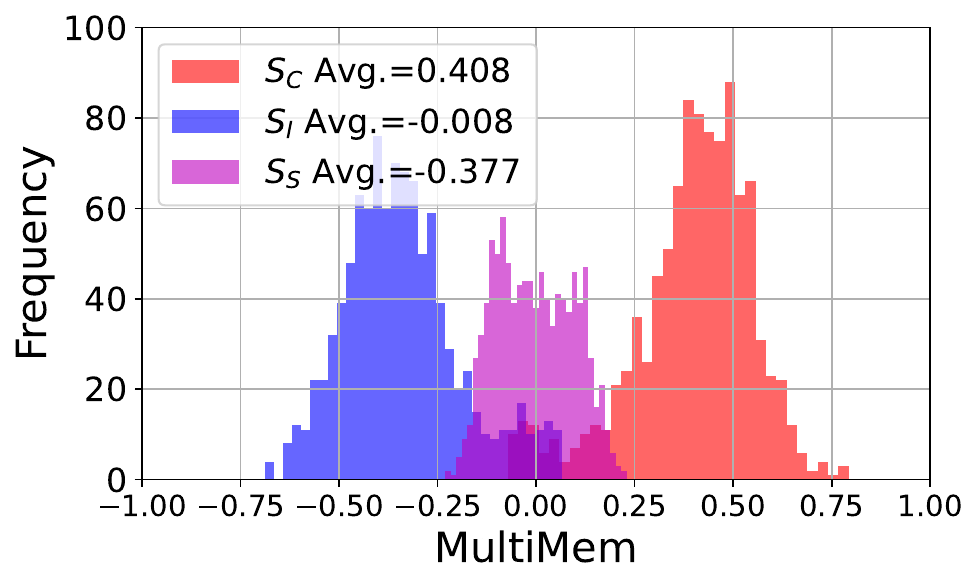}
        \caption{\ours (AVT).}
    \end{subfigure}
    \begin{subfigure}[b]{0.45\textwidth}
        \centering
        \includegraphics[width=\textwidth]{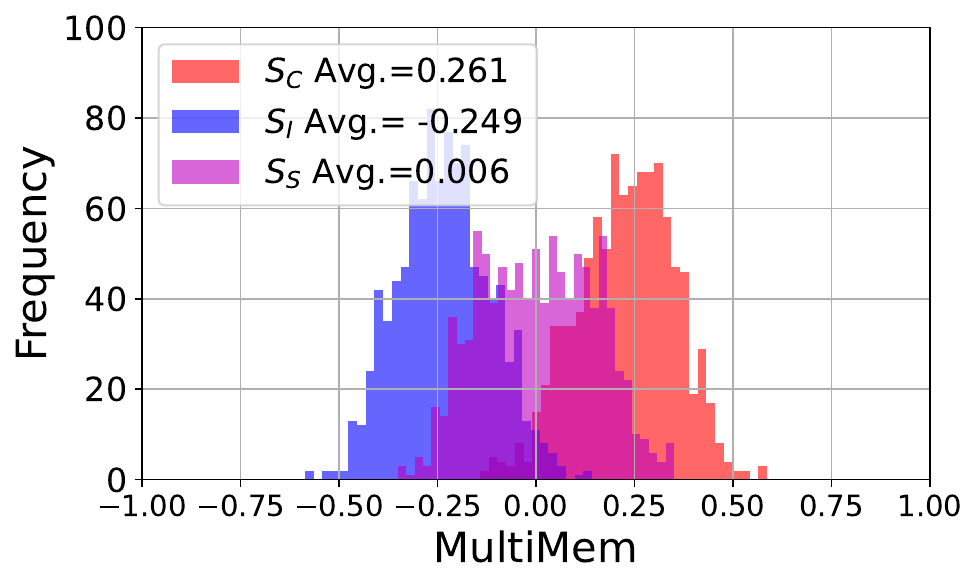}
        \caption{\ours (AV).}
    \end{subfigure}  
    \begin{subfigure}[b]{0.45\textwidth}
        \centering
        \includegraphics[width=\textwidth]{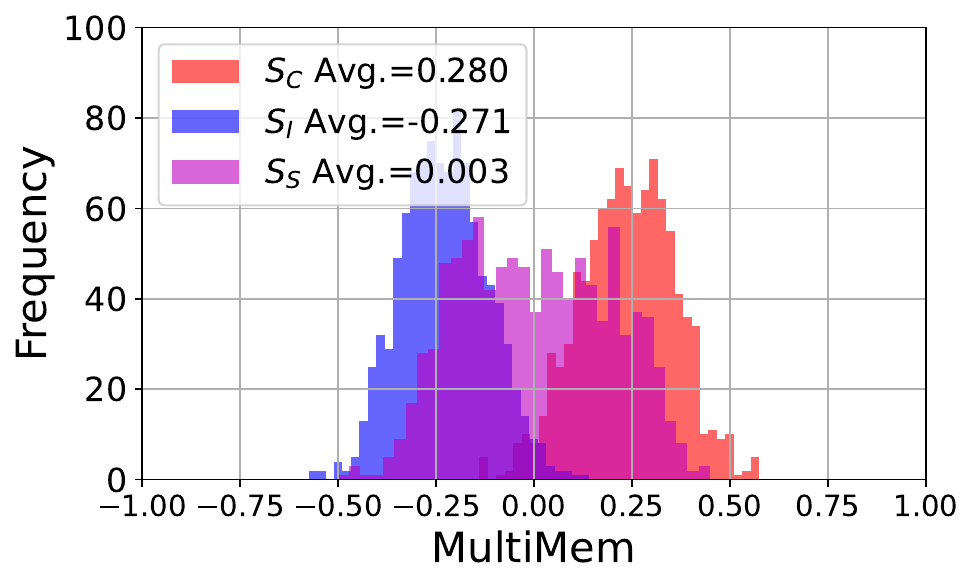}
        \caption{\ours (AT).}
    \end{subfigure}
    \begin{subfigure}[b]{0.45\textwidth}
        \centering
        \includegraphics[width=\textwidth]{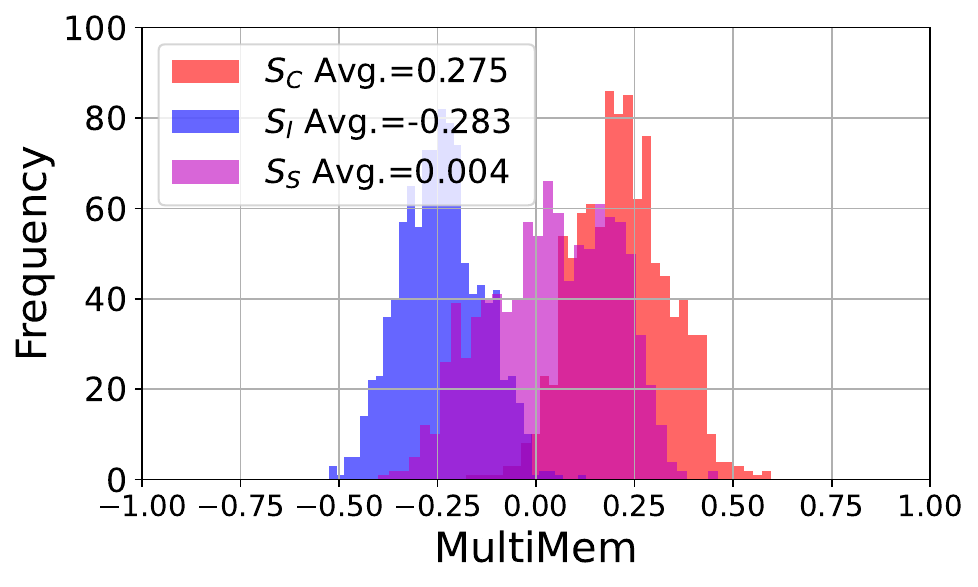}
        \caption{\ours (VT).}
    \end{subfigure}
    \caption{\textbf{Measuring memorization only on modality pair is insufficient for \avt trained on MSRVTT.} (a) Measure memorization across all three modalities (Audio, Video, and Text) of AudioCLIP. 
    (b)-(d) Measure pairwise memorization on all modality pairs (Audio-Video, Audio-Text, and Video-Text)of \avt. }
    \label{fig:all_multi_vs_clip_AVT}
\end{figure*}

\begin{figure*}[t]
    \centering
    \begin{subfigure}[b]{0.32\textwidth}
        \centering
        \includegraphics[width=\textwidth]{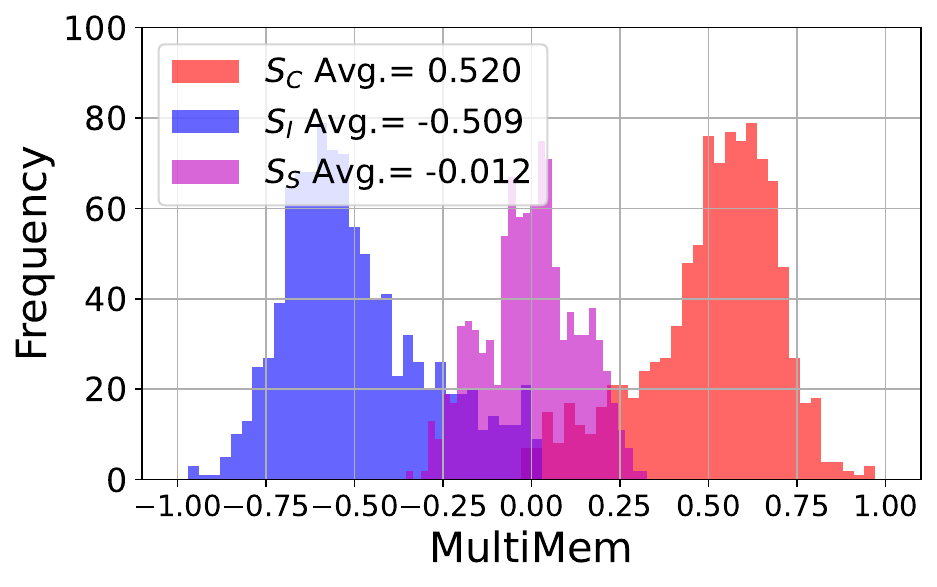}
        \caption{\ours (AVIT).}
    \end{subfigure}
    \begin{subfigure}[b]{0.32\textwidth}
        \centering
        \includegraphics[width=\textwidth]{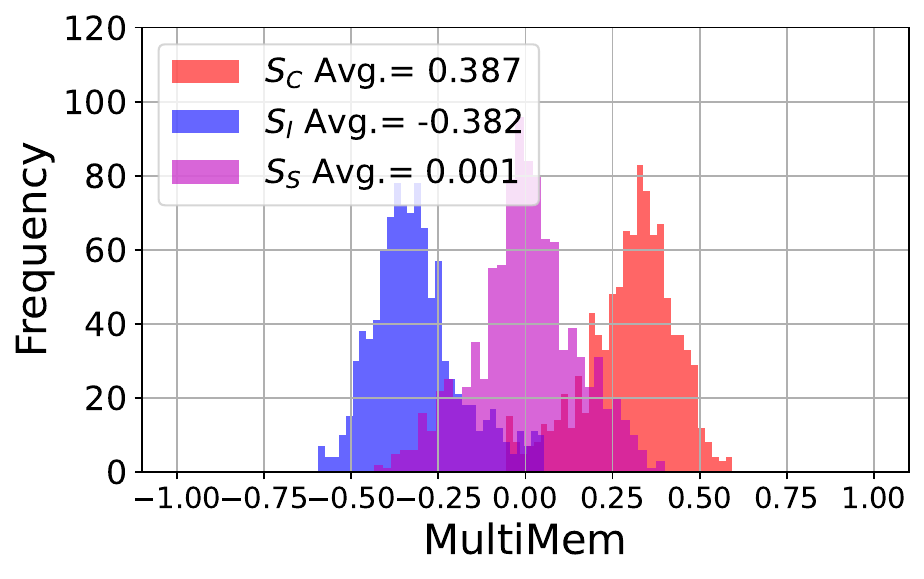}
        \caption{\ours (AVT).}
    \end{subfigure}  
    \begin{subfigure}[b]{0.32\textwidth}
        \centering
        \includegraphics[width=\textwidth]{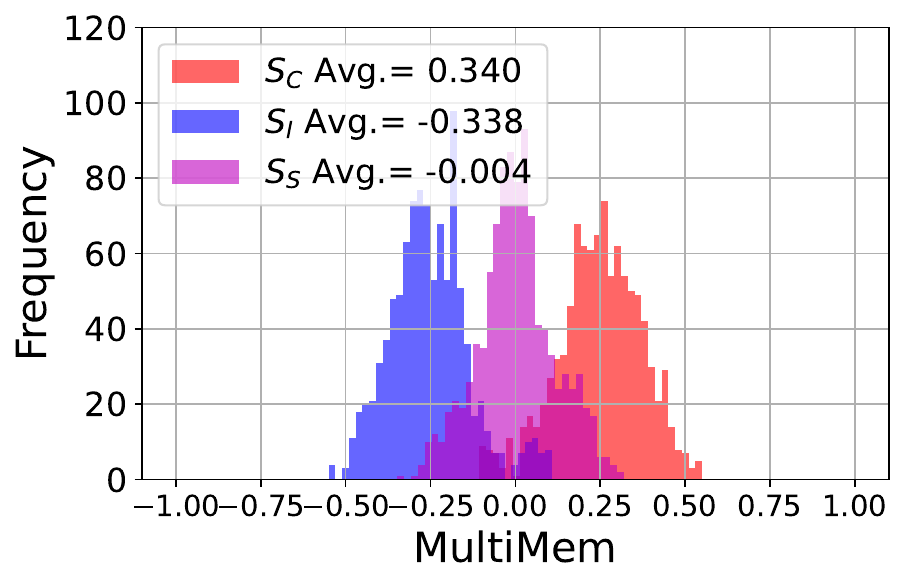}
        \caption{\ours (AIT).}
    \end{subfigure}
    \caption{\textbf{Measuring memorization only on partial modality is insufficient for \avit.} (a) Measure memorization across all four modalities (Audio, Video, Image, and Text) of AudioCLIP. 
    (b) Measure memorization with AVT \ours. (c) Measure memorization with AIT \ours.}
    \label{fig:all_multi_vs_clip_AVIT}
\end{figure*}

\paragraph{\textbf{Measuring Memorization in \avt and \avit.}}\label{app:quantify_avit}
For \avt, we measure the memorization with two metrics: 1) tri-modal \ours with all three modalities and 2) bi-modal \ours of all modality pairs (A-V, A-T, and V-T). The results are shown in \Cref{fig:all_multi_vs_clip_AVT}, which aligns with the results for AudioCLIP in the main paper.
For \avit, we also measure the memorization with two metrics: 1) quad-modal \ours with all four modalities and 2) tri-modal \ours for VideoCLIP and \avt (\ie AVT \ours and AIT \ours). The results are shown in \Cref{fig:all_multi_vs_clip_AVIT}. We can find that when \avt or AudioCLIP is extended to a quad-modal setting, the previously used tri-modal \ours becomes insufficient in capturing both the distribution of memorized samples and the accurate measurement of memorization levels. This again highlights the necessity of using all modalities to measure memorization.

\begin{figure*}[t]
    \centering
    \includegraphics[width=0.6\textwidth]{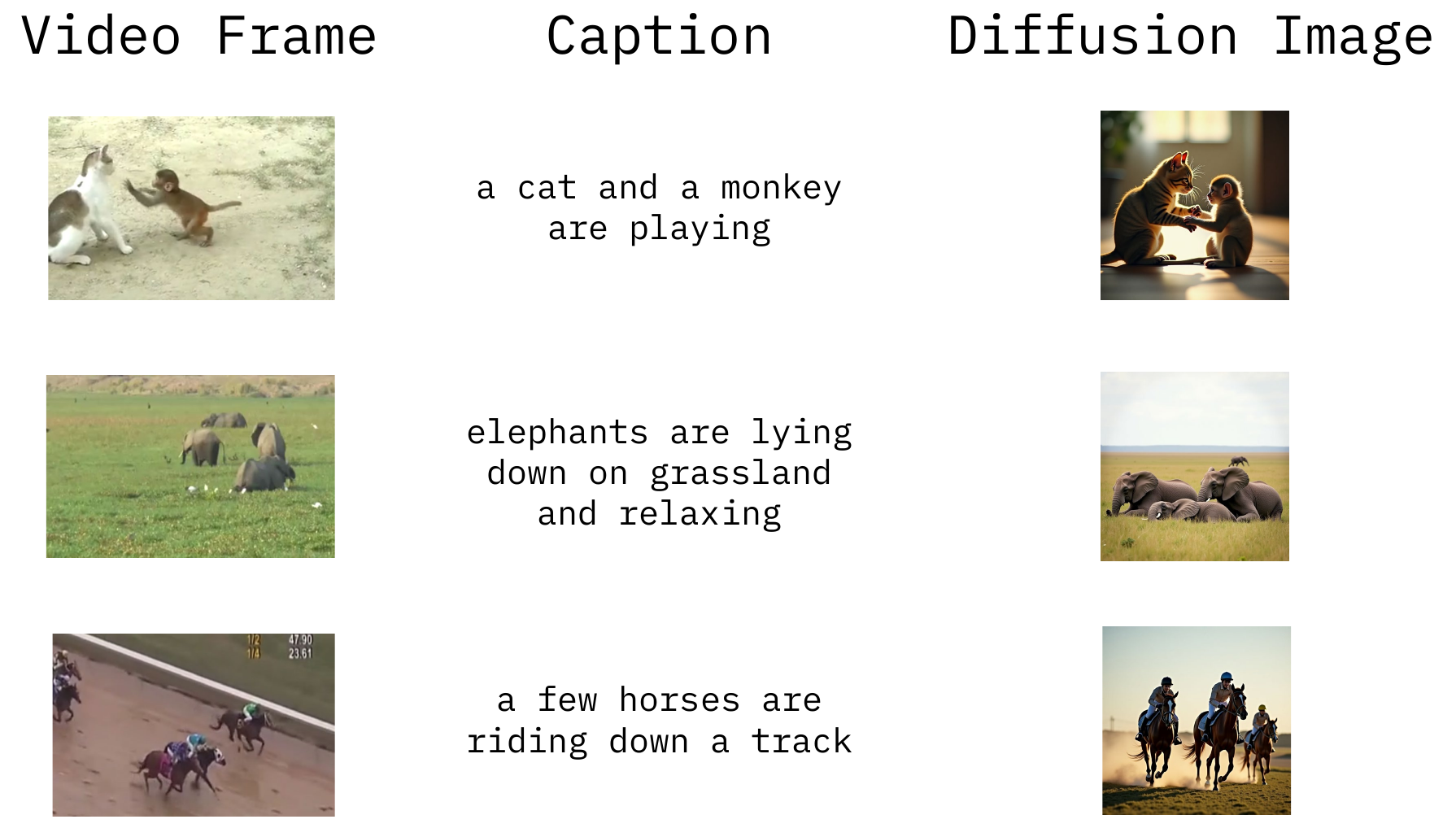}
    \caption{\textbf{Samples of images generated by Stable Diffusion v1.5.}}
    \label{fig:image_gen}
\end{figure*}

\paragraph{\textbf{Measuring Memorization \avit with synthesized images.}}\label{app:quantify_avit_gen}
We use Stable Diffusion v1.5 to generate images from the captions of the MSRVTT dataset to construct a dataset with true four modalities: (1) audio, (2) video, (3) image, and (4) text. 
We adopt this setup based on the assumption that using newly generated images with styles different from the video frames can mitigate the potential unnecessary memorization caused by overly similar visual styles between images and video frames.
Note that we synthesize the images for the MSRVTT dataset because there are no natural 4-modality datasets. A few examples of generated images are presented in \Cref{fig:image_gen}.

\begin{figure*}[t]
    \centering
    \begin{subfigure}[b]{0.32\textwidth}
        \centering
        \includegraphics[width=\textwidth]{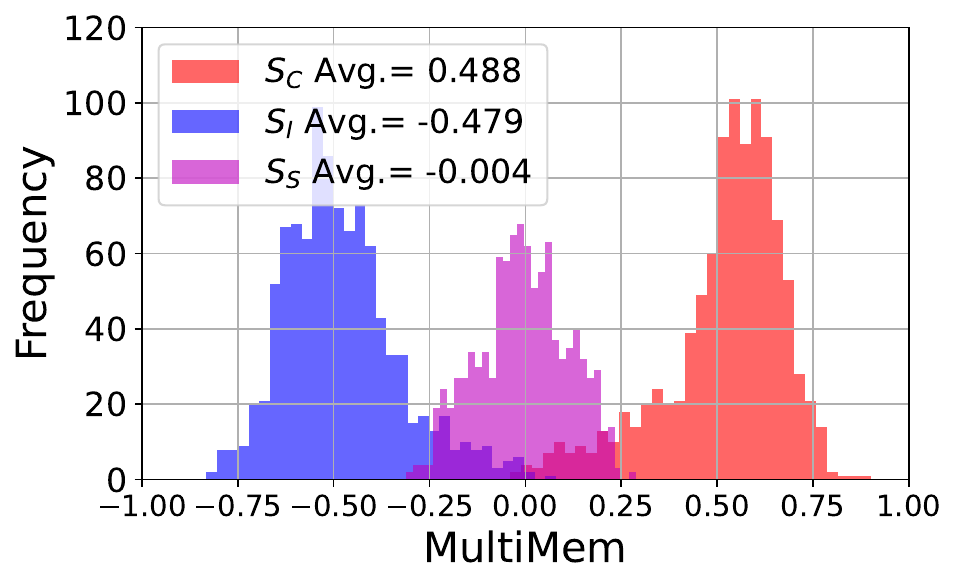}
        \caption{\ours (AVIT).}
    \end{subfigure}
    \begin{subfigure}[b]{0.32\textwidth}
        \centering
        \includegraphics[width=\textwidth]{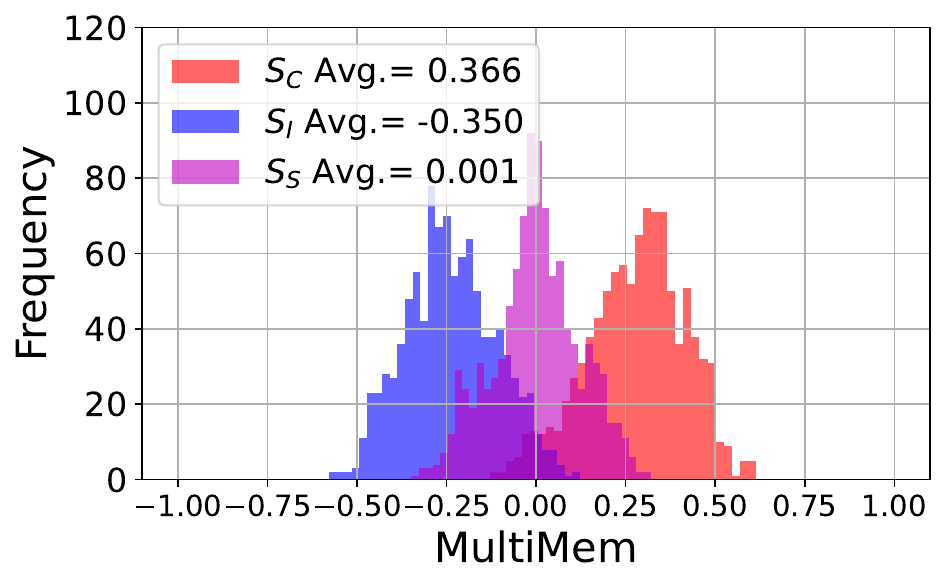}
        \caption{\ours (AVT).}
    \end{subfigure}  
    \begin{subfigure}[b]{0.32\textwidth}
        \centering
        \includegraphics[width=\textwidth]{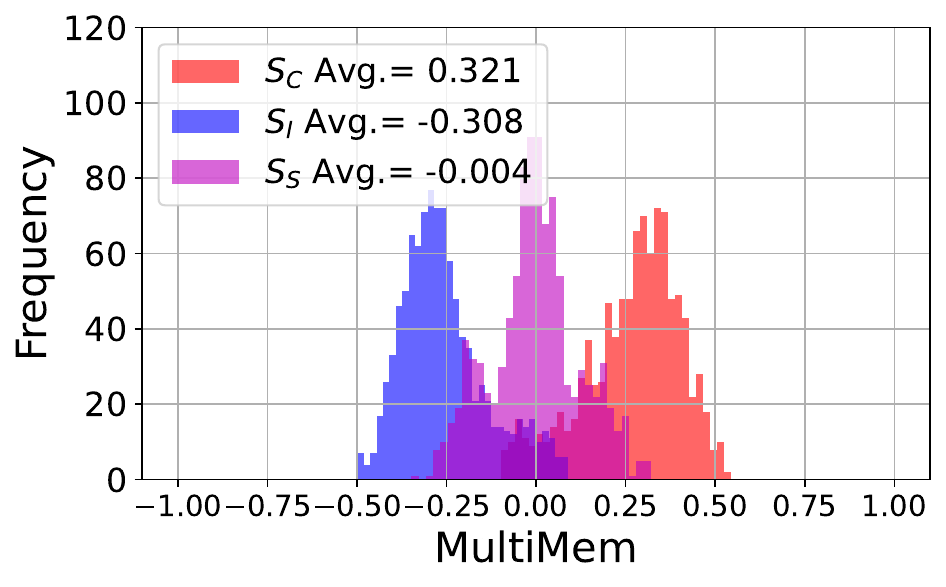}
        \caption{\ours (AIT).}
    \end{subfigure}
    \caption{\textbf{\avit trained with synthesized images follows the same trend with results trained with video-frame images.} (a) Measure memorization across all four modalities (Audio, Video, Image, and Text) of AudioCLIP. 
    (b) Measure memorization with AVT \ours. (c) Measure memorization with AIT \ours.}
    \label{fig:all_gen_multi_vs_clip_AVIT}
\end{figure*}

The results are shown in \Cref{fig:all_gen_multi_vs_clip_AVIT}, we find the memorization distribution fully follow the same trend observed in \Cref{fig:all_multi_vs_clip_AVIT}.

\paragraph{\textbf{Top 10 most memorized samples for VideoCLIP and \avt.}}\label{app:top10}
Unlike images, videos contain dynamic visual and audio signals, which cannot be effectively displayed in the paper. To address this, 
We provide two examples here and others are shown in supplementary materials.
As shown in \Cref{fig:most_mem1}, video3405 is a completely dark-screen video, providing almost no visual semantic information relevant to its caption: “a man talking about hydroponic fluid pressure.”
As a result, it ranks first in VideoCLIP with a high \ours of 0.529. 
However, its audio clearly mentions “hydroponic fluid pressure”, which aligns well with the caption. 
Therefore, its \ours score in \avt is only 0.561, ranking 36th. Another example is video6659 shown in \Cref{fig:most_mem2}, where the visual content features two animated characters, while the audio contains a conversation between two male-like voices. The title, “a character hunting for love”, is poorly related to both the video and audio. All three modalities are semantically misaligned. As a result, this sample receives a high \ours score of 0.712 in \avt, ranking second among all samples.

\begin{figure*}[t]
    \centering
    \begin{subfigure}[b]{0.45\textwidth}
        \centering
        \includegraphics[width=\textwidth]{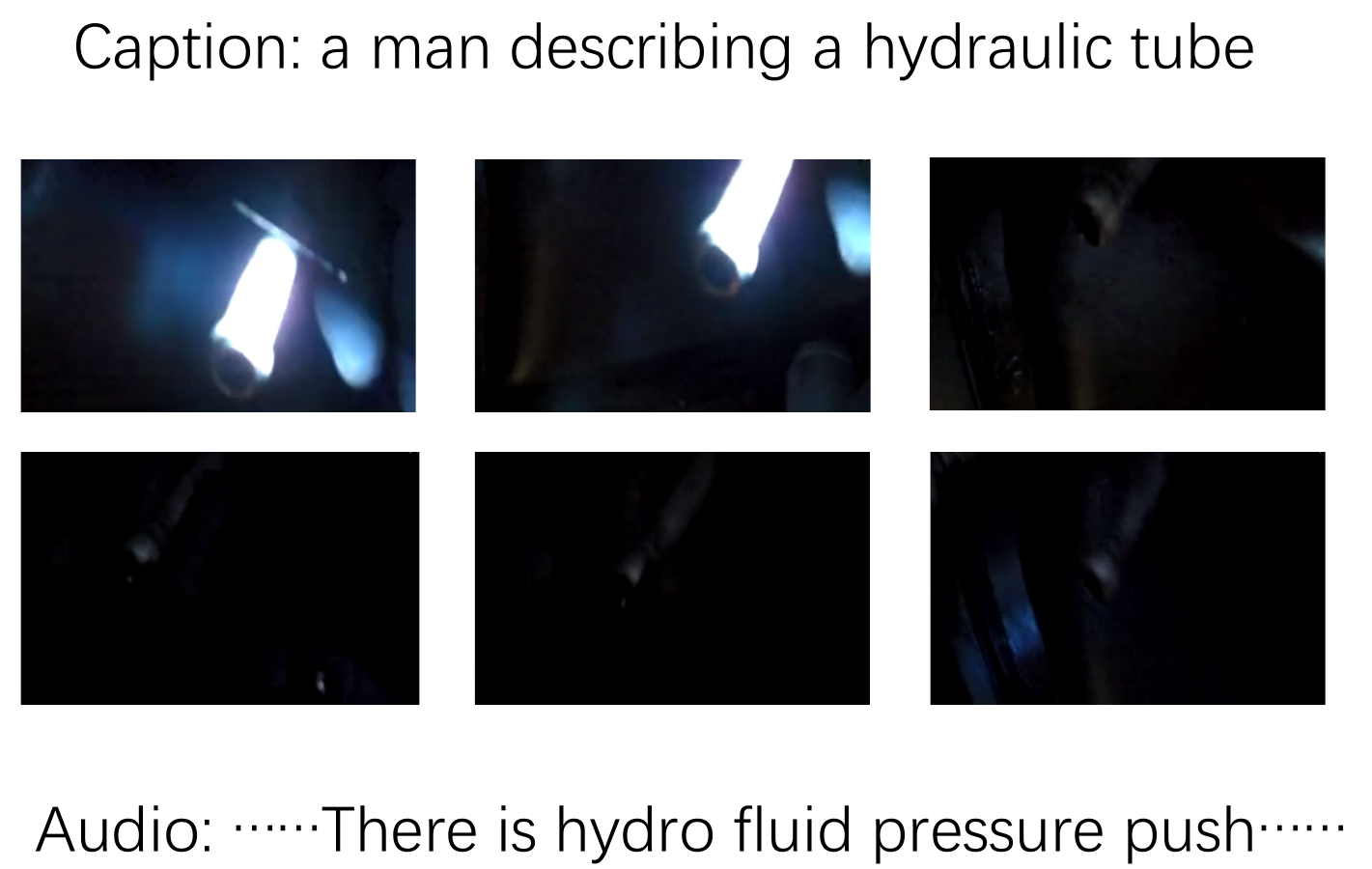}
        \caption{Video3405}
        \label{fig:most_mem1}
    \end{subfigure}
    \begin{subfigure}[b]{0.45\textwidth}
        \centering
        \includegraphics[width=\textwidth]{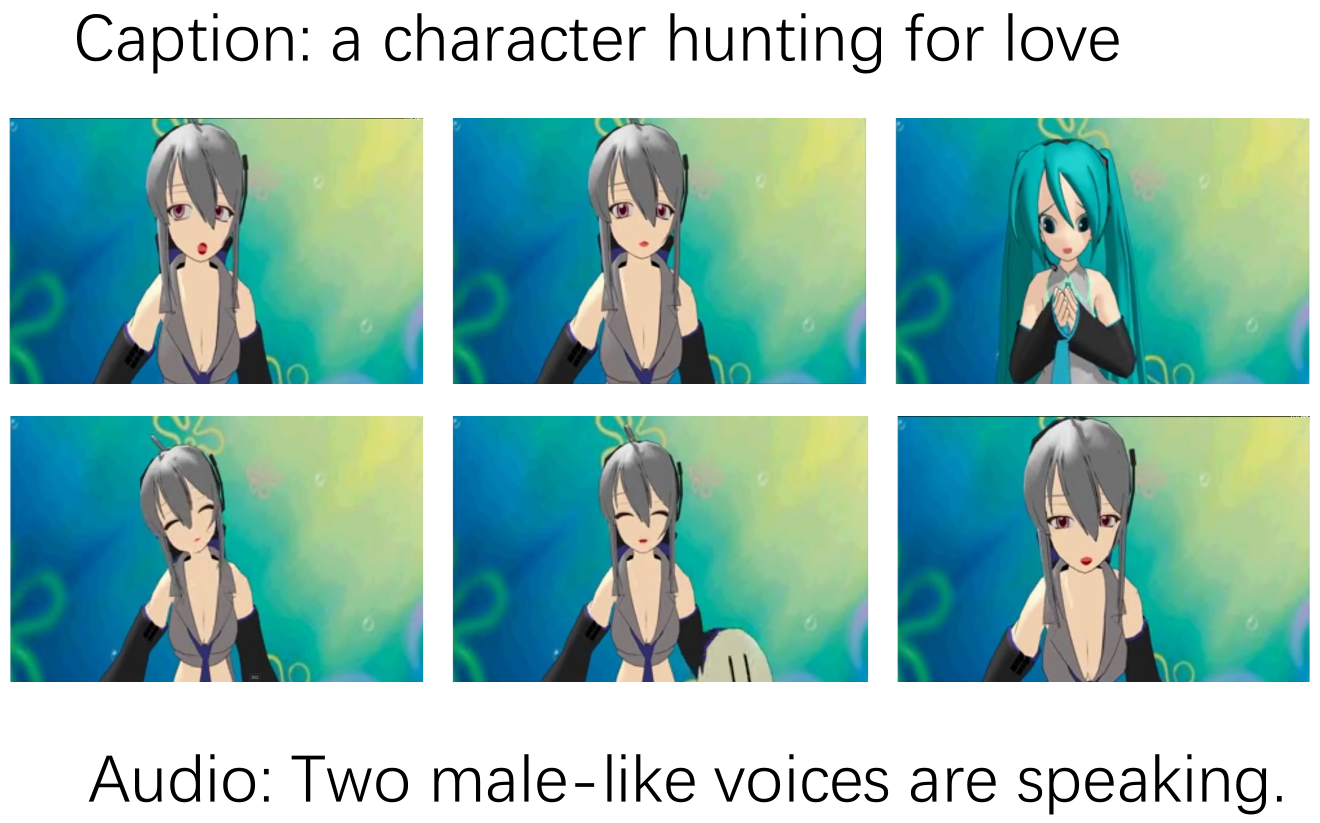}
        \caption{Video6659}
        \label{fig:most_mem2}
    \end{subfigure}  

    \caption{\textbf{The most memorized samples for VideoCLIP and \avt.}  }
    \label{fig:most_mem}
\end{figure*}

\subsection{Additional Experiments}\label{app:ex-exp}

\paragraph{\textbf{ImageBind Experiment Setup}}
We follow the original design of ImageBind that leverage the image embedding as the unified anchor space. All other modalities are aligned to this space through image-to-modality contrastive supervision. We keep the audio, video, and text encoders unchanged as random initialization, but replace the image encoder with our own pretrained image encoder. The encoder produces a normalized image representation that serves as the reference for all contrastive objectives. This design removes all non-image pairwise losses (e.g., audio–video, audio–text, and video-text), ensuring that the image modality forms the central representation space. We use Adam optimizer with a learning rate of 1e-4 and train for 200 epochs. The batch size and dataset split follow the implementation of AVIT-CLIP.

\paragraph{\textbf{Enhancing Generalization by Introducing new Modalities.}}
\label{app:intro_modality}
Our results in \Cref{tab:performance_avit} show that, when extending VideoCLIP to \avt, the accuracy of V-T retrieval task increases. This is because adding a new modality to VideoCLIP reduces the text modality's dominance in memorization, yielding a more balanced memorization distribution across modalities. So that increasing model's generalization. This is consistent with our claims in \Secref{sec:balance}. 
Similar trend could be observed in extending \avt or AudioCLIP to \avit. 

Note that the \avit always has the best performance in all retrieval tasks (as reflected by the boldface values in the \Cref{tab:performance_avit}).

\begin{table*}[t]
    \centering
    \small
    \resizebox{\textwidth}{!}{\begin{tabular}{ccccccccccccc}
    \toprule

Model&\makecell{VT\\ \ours}&\makecell{AVT\\ \ours}  & \makecell{AIT\\ \ours} &\makecell{AVIT\\ \ours} &\makecell{A\\SSLMcm}&\makecell{V\\SSLMcm}& \makecell{I\\SSLMcm}& \makecell{T\\SSLMcm} & \makecell{T@5(\%)\\V-T} & \makecell{T@5(\%)\\AV-T} &\makecell{T@5(\%)\\AI-T}&\makecell{T@5(\%)\\AVI-T}\\
    \midrule
VideoCLIP&0.347&-&-&-&-&0.179&-&0.206&32.7&-&-&-\\
\avt&0.275&0.408&-&-&0.238&0.251&-&0.266&33.9&40.4&-&-\\
AudioCLIP&-&-&0.332&-&0.188&-&0.191&0.210&-&-&36.9&-\\
Imagebind-AVIT&0.289&0.414&0.339&0.571&0.220&0.231&0.238&0.242&33.0&35.1&32.9&36.5\\
\avit&0.254&0.366&0.321&0.488&0.244&0.255&0.260&0.267&\textbf{36.6}&\textbf{41.3}&\textbf{37.0}&\textbf{48.2}\\
\bottomrule     
\end{tabular}}
\caption{\textbf{Introducing a new modality helps reduce the global memorization, and in turn, improves the model's generalization.} }
\label{tab:performance_avit}
\end{table*}

\paragraph{\textbf{Effects of Augmenting Different Ratios of Most Memorized Samples.}}\label{app:augment_ratio}

\begin{figure*}[t]
    \centering
    \includegraphics[width=0.5\textwidth]{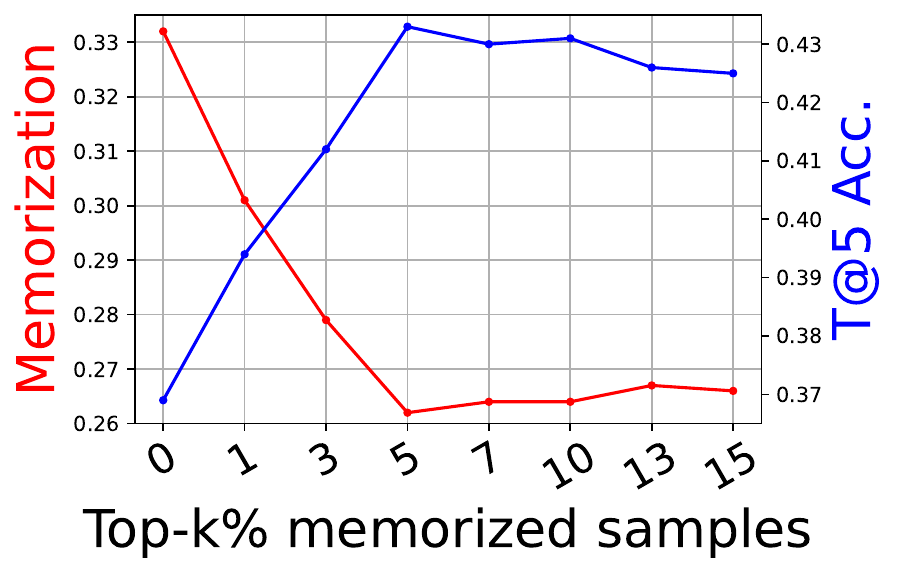}
    \caption{\textbf{Effects of noise-based augmentation to the top k\% most memorized samples (ranked by \ours) during training.} }
    \label{fig:augment_ratio}
\end{figure*}

In \Cref{fig:augment_ratio}, we test the effect of applying noise-based augmentation to different proportions of the most memorized samples during training. The results show that when noise-based augmentation is applied to less than 5\% of the most memorized samples, global memorization decreases rapidly, and overall performance improves significantly. However, once the proportion exceeds 5\%, both memorization and performance begin to stabilize as more samples are augmented. This indicates that decoupling the cross-modal associations of non-highly memorized samples is not effective in mitigating global memorization or improving performance. In contrast, applying augmentation only to the most memorized samples identified by \ours achieves the best results with significantly lower computational cost.

\paragraph{\textbf{\unit results on COCO.}}\label{app:unitmem_coco}

In this experiment, we use the original COCO dataset as the source of both image inputs and text captions. Speeches generated from COCO captions by open-source Tortoise TTS are used as audio modality. For SL, the model is trained using a multi-label classification setting with the cross-entropy loss with COCO dataset. All other settings for SSL, CLIP and AudioCLIP are similar to those in the experiments with UrbanSound8K dataset.
We adopt this setup based on the assumption that captions and the audio generated from them share higher cross-modal semantic alignment. This helps mitigate the potential impact of weak semantic alignment among spectrograms, audio, and textual labels in the UrbanSound8K dataset. 
Note that the synthesized audio for COCO only used as 
extended evidence to verify the trend observed on the UrbanSound8K dataset in main paper.

\begin{figure*}[t]
    \centering
    \includegraphics[width=0.6\textwidth]{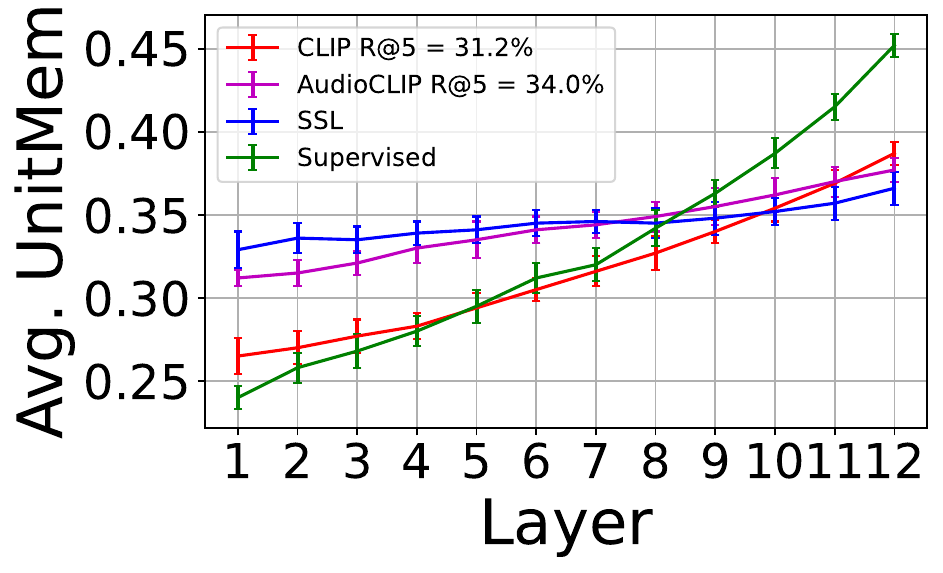}
    \caption{\textbf{\unit: AudioCLIP is more aligned with SSL.} We implement an extra experiment on COCO with audios generated from the captions.}
    \label{fig:image_coco_unitmem}
\end{figure*}

The results are shown in \Cref{fig:image_coco_unitmem}. 
Despite semantic alignment across modalities being strengthened, we observe that the behavior of AudioCLIP’s vision encoder remains more similar to SSL than to SL. This fully aligns with the result of that implemented on UrbanSound8K dataset in main paper. 

\begin{table*}[t]
    \centering
    \begin{tabular}{ccc}
    \toprule
Regrouping epoch &\ours& AI-T retrieval T@5\\
\midrule
None&0.332&36.9\\
10&0.321&38.2\\
80&0.262&43.3\\
10 + 80&0.256&43.7\\
20 + 80&0.252&44.1\\
10 + 90&0.254&43.8\\
40 + 50&0.266&42.8\\
20 + 50 + 80&0.250&44.2\\
10 + 40 + 70&0.253&43.8\\

\bottomrule     
\end{tabular}
\caption{\textbf{Applying the \textit{in-training regrouping} strategy multiple times further mitigate memorization and improves performance.} }
\label{tab:twice}
\end{table*}

\paragraph{\textbf{Mitigating Memorization During Training.}}\label{app:mit_mem_intrain}

In addition to the experiments presented in Section “\textbf{In-training Mitigation}”, we implement three additional experiments to further validate the effectiveness of this approach. 

\begin{figure*}[t]
    \centering
    \includegraphics[width=0.6\textwidth]{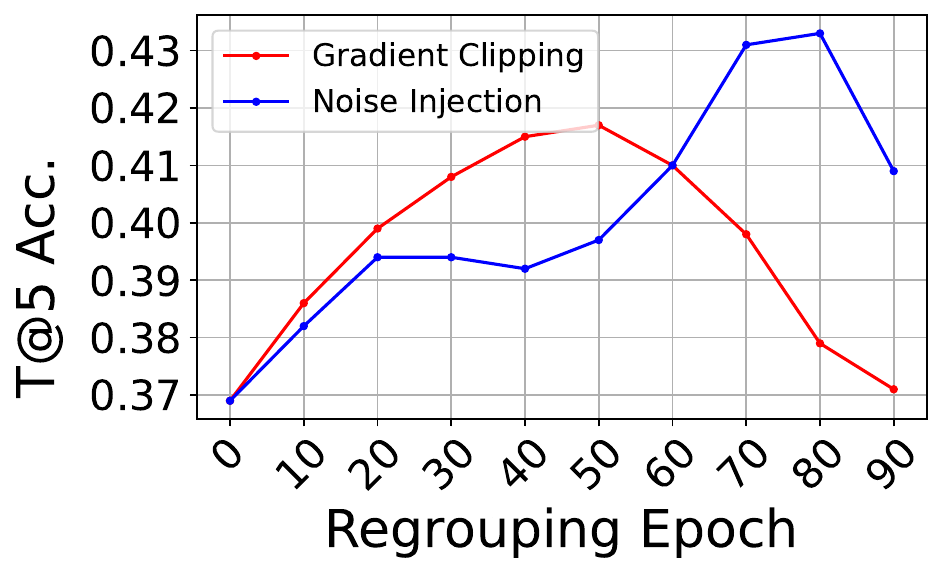}
    \caption{\textbf{Performance for AudioCLIP under in-training regrouping when using noise-based augmentation and gradient clipping.}}
    \label{fig:nj_gc}
\end{figure*}

In the first experiment, we compared the different effects of noise-based augmentation and gradient clipping during the in-training mitigation process. The results in \Cref{fig:nj_gc} shows that gradient clipping needs a much larger learning rate than noise-injection to mitigate the memorization level of the model. Besides, the best performance of gradient clipping (41.7\% +4.8\% to 36.9\% baseline) is worse than noise-injection(43.3\% +6.7\% to 36.9\% baseline)

\begin{figure*}[t]
    \centering
    \includegraphics[width=0.6\textwidth]{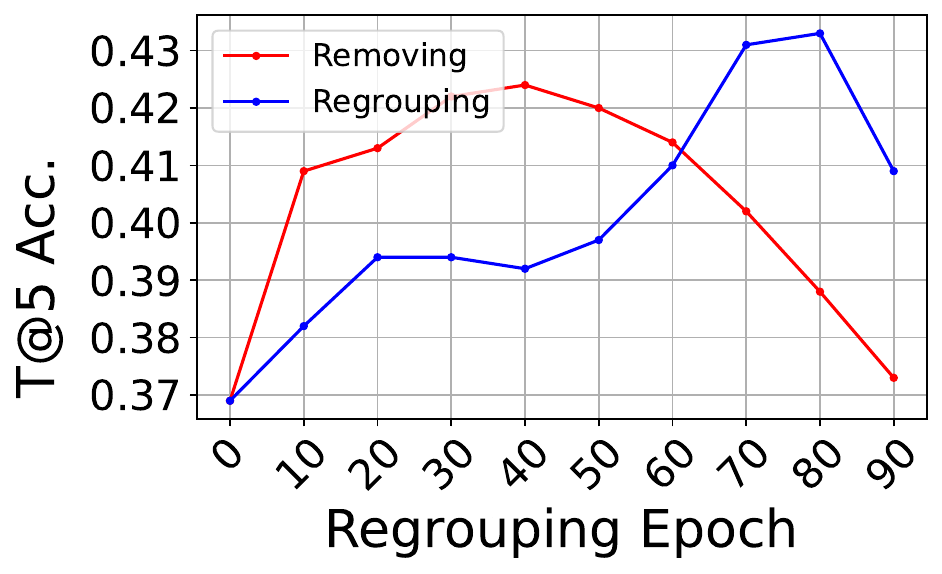}
    \caption{\textbf{Performance for AudioCLIP under in-training regrouping vs. in-training removal.}}
    \label{fig:regroup_remove}
\end{figure*}

In the second experiment, we still use \ours to measure the memorization of all training samples at every 10-epoch interval. However, instead of regrouping the top 5\% most memorized samples into new batches with noise-based augmentation, we directly \textbf{remove} them from the training set. 
As shown in \Cref{fig:regroup_remove}, the performance of \textit{in-training removal} first increases and then drops. This performance drop may be due to similar reasons observed in \textit{in-training regrouping},\ie the learning rate becomes too low in the later stages of training, and the remaining number of epochs is insufficient to fully eliminate the negative impact introduced by previously highly memorized samples. Moreover, we observe that the maximum performance gain achieved by  \textit{in-training removal} is smaller than that of \textit{in-training regrouping}, further demonstrating the effectiveness of our proposed strategy.

In the third experiment, we further examine that whether applying the \textit{in-training regrouping} strategy multiple times will provide additional memorization mitigation and performance improvement or not.
Specifically, we apply the \textit{in-training regrouping} strategy twice: once at the end of the warm-up phase (epoch 10) and once near the end of training (epoch 80). The results in \Cref{tab:twice} show that applying the \textit{in-training regrouping} strategy multiple times can further mitigate model memorization and yield better generalization. However, using it three times can hardly achieve a significant performance improvement compared to using it twice, so using it twice at the beginning and end of the training is the best choice considering that it does not bring many extra calculations.


\paragraph{Model Size and Dataset Size.} 
We added additional experiments on CLIP models of varying sizes trained on the same data, as well as models of the same size trained on different amount of the training data. The results in~\Cref{tab:size} show that larger models exhibit higher memorization levels when trained on the same data, and also reducing the amount of training data leads to increased memorization for models of the same size. Both trends are consistent with prior work (e.g., [Carlini et al., ICLR 2022], [Tirumala et al., NeurIPS 2022], [27]), as larger model capacity provides more room to retain instance-level details, while sparser data coverage reduces generalization opportunities and increases reliance on memorizing individual samples.

\begin{table*}[t]
    \centering
    \begin{adjustbox}{width=\textwidth}
\begin{tabular}{lcccccc}
    \toprule
&Baseline& ViT-S + EsResNet-34 & ViT-L + EsResNet-101 &75\% dataset usage&50\% dataset usage& 25\% dataset usage\\
\midrule
Avg. MultiMem&0.332&0.308&0.382&0.341&0.360&0.389\\
    \bottomrule     
\end{tabular}
\end{adjustbox}
\caption{\textbf{Memorization v.s. model size and training datapoint usage.} }
\label{tab:size}
\end{table*}

\end{document}